%% file: main.tex
\documentclass{bmvc2k}

\usepackage{color,array}
\usepackage{tikz}
\usepackage{graphicx}
\usepackage{amsmath,amssymb} 
\usepackage{siunitx}
\usepackage[normalem]{ulem}
\usepackage{slashbox}
\usepackage{subfigure}
\usepackage{epsfig}
\usepackage{multirow}
\usepackage{times}
\usepackage{url}
\usepackage{caption}
\usepackage{dirtytalk}
\usepackage{mathtools}
\usepackage{booktabs}
\usepackage{algorithm}
\usepackage[noend]{algpseudocode}
\usepackage{bbm}
\usepackage{xspace}
\usepackage{hyperref}  
\usepackage{capt-of}
\usepackage{subfigure}
\usepackage{balance}

\usepackage{hhline,colortbl}
\usepackage{makecell}
\usepackage{comment}

\graphicspath{{./}{./figures/}}
\DeclareGraphicsExtensions{.pdf,.jpeg,.png,.jpg,.eps}
\usepackage{caption}
\usepackage[misc]{ifsym}

\DeclareRobustCommand\onedot{\futurelet\@let@token\@onedot}
\def\onedot{\ifx\@let@token.\else.\null\fi\xspace}

\definecolor{Gray}{gray}{0.9}
\definecolor{Gray1}{gray}{0.97}
\graphicspath{{./}{./figures/}}
\DeclareGraphicsExtensions{.pdf,.jpeg,.png,.jpg,.eps}

\definecolor{ao}{rgb}{0.0, 0.5, 0.0}
\def\onedot{\ifx\@let@token.\else.\null\fi\xspace}
\def\eg{\emph{e.g}\onedot} 
\def\eg{\emph{e.g}\onedot} 
\def\ie{\emph{i.e}\onedot}

\def\etal{\emph{et al}\onedot}
\makeatother

\def\Vec#1{{\boldsymbol{#1}}}
\def\Mat#1{{\boldsymbol{#1}}}

\def\C#1{{\color{black} {#1}}}
\def\D#1{{\color{black} {#1}}}

\newtheorem{remark}{Remark}

\newcommand{\algrule}[1][.2pt]{\par\vskip.5\baselineskip\hrule height #1\par\vskip.5\baselineskip}

\newcommand{\tr}{\mathop{\rm  Tr}\nolimits}

\newcommand{\RNum}[1]{\lowercase\expandafter{\romannumeral #1\relax}}
\def\thefootnote{*}\footnotetext{These authors contributed equally to this work.}

\newlength{\Oldarrayrulewidth}

\usepackage{float}

\def\main{1} 

\if\main1
    \title{Learning Deep Optimal Embeddings with Sinkhorn Divergences}
\else
    \title{Supplementary: Learning Deep Optimal Embeddings with Sinkhorn Divergences}
\fi
\addauthor{Soumava Kumar Roy\thefootnote{}}{soumava.roy@epfl.ch}{1}
\addauthor{Yan Han\thefootnote{}}{Yan.Han@anu.edu.au}{2,4}
\addauthor{Mehrtash Harandi}{mehrtash.harandi@monash.edu}{3}
\addauthor{Lars Petersson}{Lars.Petersson@data61.csiro.au}{4}

\addinstitution{
 CVLAB EPFL, Switzerland
}
\addinstitution{
Australian National University, \\
Canberra, Australia
}

\addinstitution{Monash University \\
Melbourne, Australia}

\addinstitution{Data61/CSIRO, \\
Canberra, Australia
}

\runninghead{Soumava, Yan, Mehrtash, Lars}{Deep Class-wise Discrepancy Loss}

\def\eg{\emph{e.g}\bmvaOneDot}

\def\etal{\emph{et al}\bmvaOneDot}

\begin{document}
\maketitle
        \input{SOUMAVA-FILES/paper_abstract_ivc}
        \input{SOUMAVA-FILES/paper_introduction}

\input{SOUMAVA-FILES/paper_methodology}
        \input{SOUMAVA-FILES/paper_related}
        \input{SOUMAVA-FILES/paper_simple_exps}
        \input{SOUMAVA-FILES/paper_conclusion}

    \newpage
    {
     	\centering
    	\textbf{\LARGE Learning Deep Optimal Embeddings with Sinkhorn Divergences ---	Supplementary Information\\}
    	\vspace{0.5 em}
    	}
        \input{SOUMAVA-FILES/paper_prelims}
        \input{SOUMAVA-FILES/paper_ablation}

\input{SOUMAVA-FILES/paper_clean_labels}

\input{SOUMAVA-FILES/paper_fgor}
\bibliography{references}
\end{document}

%% file: SOUMAVA-FILES/paper_abstract_ivc.tex
\begin{abstract}
        Deep Metric Learning algorithms aim to learn an efficient embedding space to preserve the similarity relationships among the input data. Whilst these algorithms have achieved significant performance gains across a wide plethora of tasks, they have also failed to consider and increase comprehensive similarity constraints; thus learning a sub-optimal metric in the embedding space. Moreover, up until now; there have been few studies with respect to their performance in the presence of noisy labels. Here, we address the concern of learning a discriminative deep embedding space by designing a novel, yet effective Deep Class-wise Discrepancy Loss (DCDL) function that segregates the underlying similarity distributions (thus introducing class-wise discrepancy) of the embedding points between each and every class. Our empirical results across three standard image classification datasets and two fine-grained image recognition datasets in the \textbf{presence} and \textbf{absence} of noise clearly demonstrate the need for incorporating such class-wise similarity relationships along with traditional algorithms while learning a discriminative embedding space. 
        \end{abstract}

%% file: SOUMAVA-FILES/paper_introduction.tex
\section{Introduction}

In this paper, we address the issue of learning an efficient, effective and discriminative non-linear embedding space that results in compact class-specific clusters of the embedded points. Towards this aim, we propose and develop a novel loss function that comprehensively incorporates \emph{class-wise} similarity attributes of the embedded points to simultaneously minimize and maximize the intra-class and inter-class variances respectively.

In general, metric learning algorithms aim to learn a parametric-functional representation of a distance metric $\Mat{M} \succeq \Mat{0}$~\citep{weinberger2009distance} to obtain distinct and compact clusters in the embedding space. Traditional algorithms either attempt to learn the feature extractors or $\Mat{M}$ separately, thus neglecting the need for learning both holistically. Interested readers are referred to~\citep{bellet2013survey,kulis2013metric} for more detailed insights into such algorithms. Deep Learning algorithms, boosted by 
\newpage 
\noindent their recent remarkable success across a variety of research areas in computer vision and machine learning, have been proposed to overcome these limitations of the traditional algorithms. These algorithms can be broadly categorized into \textbf{(a) Structure} based methods~\citep{Schroff_CVPR_2015,Song_CVPR_2017,Law_ICML_2017,wang2017deep,roy2019siamese} which preserve the local structure of the learnt embedding space, \textbf{(b) Sampling} based methods~\citep{Sohn_NIPS_2016,Wu_ICCV_2017,duan2019deep} to efficiently sample informative samples, 
\textbf{(c) Statistical} based methods~\citep{Ustinova_NIPS_2016,kumar2016learning,roth2019mic} and \textbf{(d) Generative} modeling based methods~\citep{lin2018deep,duan2018deep,zheng2019hardness} to explicitly model the intra-class variances and reduce the inter-class variances.

Irrespective of their immense successes recently, these \emph{local} deep metric learning algorithms suffer from two serious drawbacks. First, they only incorporate and maintain local geometrical or statistical constraints in their respective algorithms; thereby failing to integrate any comprehensive geometrical or statistical characteristics of the embedding space. Second, these algorithms~\citep{Ustinova_NIPS_2016,kumar2016learning} do not encapsulate \emph{class-wise} similarity/distance relationships between the embedded points, thereby failing to learn a superior discriminative embedding space to enforce the intra-class and inter-class separation constraints. A majority of the aforementioned algorithms explicitly consider class information by either pre-training~\citep{Schroff_CVPR_2015,Song_CVPR_2017,Law_ICML_2017,wang2017deep} or simultaneously training~\citep{Pengfei_BAT_NET_iccv_19,Varior_ECCV_2016} the network with \emph{partially global} cross entropy loss. This subtle, yet important practice aims to assist the network in finding a superior starting point in the loss landscape~\citep{li2018visualizing} such that an optimal solution can be attained within a minimal number of descent steps. Moreover, Horiguchi~\etal~\citep{horiguchi2019significance} also successfully demonstrated the need of employing a softmax function along with learning a discriminative metric. Even though cross-entropy loss is able to obtain class separating sub-spaces; it cannot guarantee that the resultant embedding points will be tightly knit within each of the sub-spaces. Generative modeling based algorithms~\citep{lin2018deep,zheng2019hardness} attempt to comprehensively encompass the entire embedding space by generating hard negatives~\citep{Schroff_CVPR_2015} for every embedded point. However, such generation also limits their applicability as these hard negatives need to be within the $\epsilon$ neighbourhood of the anchor; thus limiting their overall outreach in the embedding space. Distribution based statistical methods~\citep{Ustinova_NIPS_2016,kumar2016learning} fail to consider such class-wise relationships and enforce a weak disparity measure between the pairwise distributions; thereby limiting their capacity to form well separated class-wise clusters.

Keeping these drawbacks in mind, we propose and design a novel loss function known as Deep Class-wise Discrepancy Loss ($\mathrm{DCDL}$) which takes into account the \emph{class-wise} similarity relationships between the embedded points. Specifically, we compute the probability distributions of the embeddings that belong as well as do not belong to a particular class (thus \emph{class-wise}) and enforce a discrepancy constraint between the two distributions, then we \textbf{push them away} from each other with the aim to increase/decrease the intra/inter class similarities in the embedding space simultaneously. Unlike the algorithms mentioned before that do not contemplate such similarity constraints in their loss functions; $\mathrm{DCDL}$ attempts to cluster the class-wise embeddings within certain specific regions of the embedding space. Thus, $\mathrm{DCDL}$ learns to maximize the separation between the embedded points belonging to each and every distinct class while preserving their class-wise relationships. Towards achieving this objective, we employ Optimal Transport~\citep{genevay2018learning} based Sinkhorn Divergences~\citep{cuturi2013sinkhorn} to enforce discrepancy constraints between the probability distributions of the class-wise embeddings. 

The major contributions are as follows:
\begin{enumerate}
    \item A novel loss function that exploits the class-wise pairwise similarity relationships to learn a discriminative embedding space.\
    \item An augmentation to the conventional \emph{local} metric learning loss functions to learn and enforce similarity constraints in the \textbf{presence} and \textbf{absence} of noisy labels.
    \item We further show that the proposed loss function can be combined with the cross-entropy loss to further boost the overall performance of the network.
\end{enumerate}

We evaluate our proposed loss function $\mathrm{DCDL}$ on the Cifar-$10$~\citep{krizhevsky2009learning}, Cifar-$100$~\citep{krizhevsky2009learning} and STL-$10$~\citep{coates2011analysis} datasets in the \textbf{presence} and \textbf{absence} of noisy labels. We also evaluate $\mathrm{DCDL}$ on \emph{Caltech-UCSD Birds} (CUBS-200-2011)~\citep{CUB200_DB} and \emph{Stanford Cars} (CARS196) \citep{CARS196_DB} datasets in the Deep Metric Embedding Learning (DMEL) framework in both the noisy and clean labels settings. Our empirical evaluations of $\mathrm{DCDL}$ outperform the \emph{local} deep-metric learning algorithms in both noisy and clean label settings across the various datasets; which undoubtedly demonstrate the generic need of exploiting and learning such class-wise discrepancy constraints in order to learn an effective and discriminative embedding metric space.

\noindent\textbf{Notation:} Throughout this paper, we use bold lower-case letters (\eg, $\Vec{x}$) and bold upper-case letters (\eg, $\Mat{X}$) to represent column vectors and matrices respectively. $[\Vec{x}]_i$ denote the i$^{th}$ element of the vector $\Vec{x}$. $\mathbf{I}_n$ represents the ${n \times n}$ identity matrix. $\|\Mat{X}\|_F = \sqrt{\tr(\Mat{X}^\top\Mat{X})}$ denote the Frobenius norm of the matrix $\Mat{X}$, with $\tr(\cdot)$ indicating the trace of the matrix $\Mat{X}$. $\omega(\Vec{r})$ represents a diagonal matrix with diagonal elements as $\Vec{r}$. $\Mat{X}^\top$ denotes the transpose of $\Mat{X}$. $\mathcal{U}$ and $\mathcal{V}$ represent two distinct probability distributions on the metric space $\mathcal{M}$. 
$\{\Vec{u}_i\}_{i=1}^n$ and $\{\Vec{v}_j\}_{j=1}^m$ denote $n$ and $m$ i.i.d samples (or \textbf{support points}) drawn from $\mathcal{U}$ and $\mathcal{V}$ respectively.

%% file: SOUMAVA-FILES/paper_methodology.tex
\begin{figure*}[!t]
    \centering
    \includegraphics[trim=0 0cm 0cm 0cm clip,width=0.85\linewidth,keepaspectratio]{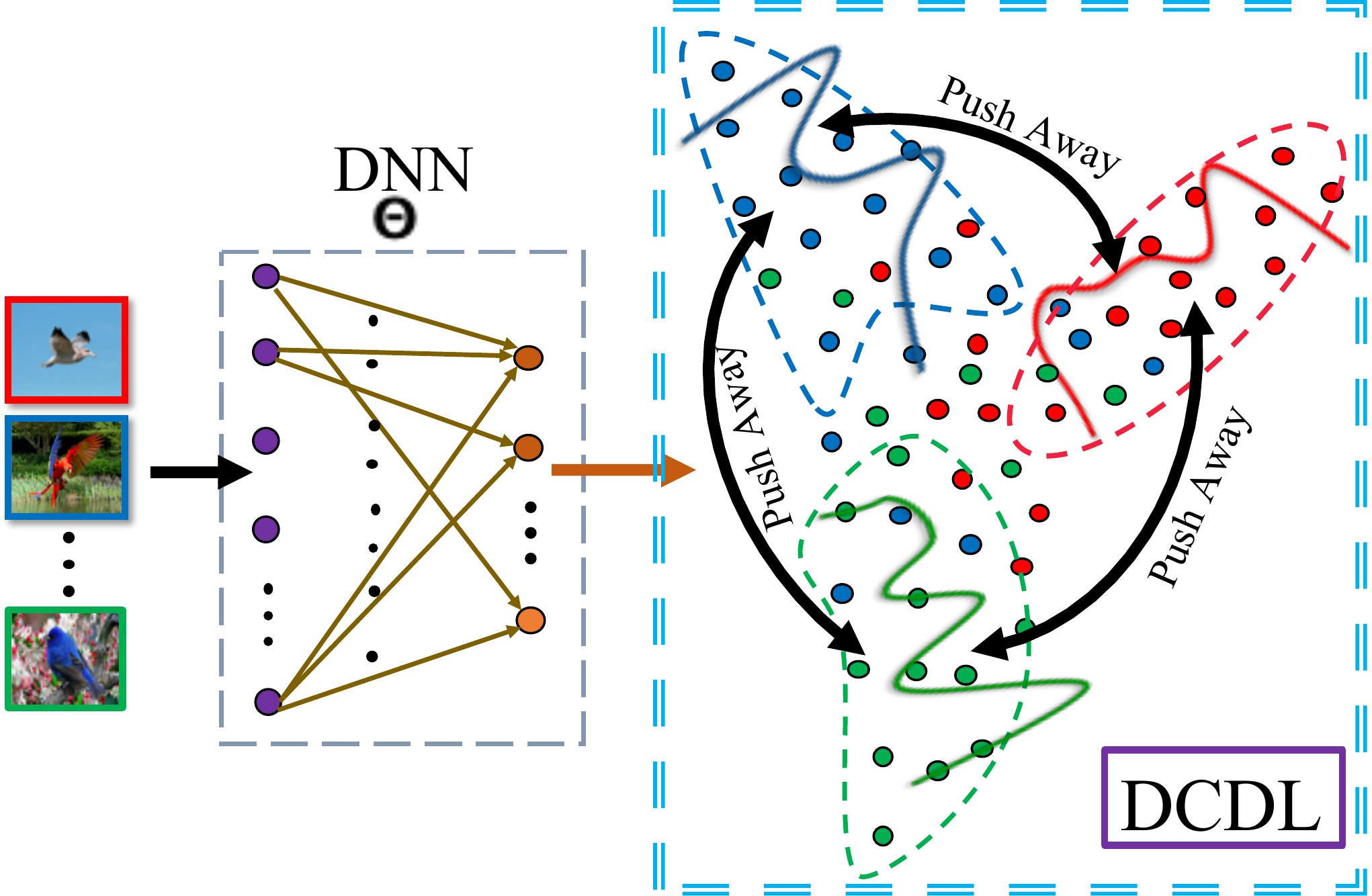}
    \vspace{0.2cm}
    \caption{\small A schematic diagram of our proposed algorithm $\mathrm{DCDL}$ is shown here. 
    }
    \label{fig:ours_DCDL}
    \vspace{-0.5cm}
\end{figure*}

\section{Deep Class-Wise Discrepancy Learning}
\label{sec:eccv2020_DCDL}

Let $\Mat{X}_i$ represent an image within the image space $\mathcal{X} \subset \mathbb{R}^{\textrm{H}\times\textrm{W}\times\textrm{C}}$. $\textrm{H}, \textrm{W}$ and $\textrm{C}$ denote the number of rows, columns and channels of $\Mat{X}_i$ respectively. The corresponding label of $\Mat{X}_i$ in the label space $\mathcal{Y} \in \left \{1, 2, ..., \textrm{K}\right \}$ is represented by $\Vec{y}_i$, where $\textrm{K}$ denotes the total number of classes in $\mathcal{X}$. The mini-batch $\mathrm{B}$ of size $\textrm{N}_{\mathrm{B}}$ is represented as $\left \{ \big(\Mat{X}_i, \Mat{y}_i\big) \right \}_{i=1}^{\textrm{N}_{\mathrm{B}}}$. In order to avoid any degenerate solutions, we ensure that there are at least $r$ samples belonging to each class within every mini-batch. A Deep Neural Network with its learnable parameters $\Theta$ is represented as $\mathcal{F}_{\Theta}$, and learns a generic non-linear mapping from $\mathcal{X}$ onto a latent feature space $\mathcal{Z} \in \mathbb{R}^{n}$; \ie $\Mat{z}_i=\mathcal{F}_{\Theta}(\Mat{X}_i)$. Moreover, the images belonging to a particular class label $k \in \mathcal{Y}$ are denoted as $\left [ \Mat{X} \right ]_k = \left \{ \Vec{X}_i \mid \Vec{y}_i = k \right \}$. Furthermore, the $\ell_2$ norm of every embedded point is constrained to be $1$ (\ie $\left \| \Mat{z}_i \right \|_2 = 1~\forall i = 1 \cdots N$). The aim of any deep embedding learning algorithm is to learn $\Theta$ such that the resulting embeddings have lower (higher) intra (inter) class variances respectively.

\noindent In the recent past, most deep learning approaches either (\RNum{1}) design a new objective function in the embedding space~\citep{Song_CVPR_2017,Law_ICML_2017,Wu_ICCV_2017}, (\RNum{2}) design efficient batch sampling techniques~\citep{Song_CVPR_2016,Sohn_NIPS_2016}, or (\RNum{3}) use generative models~\citep{lin2018deep,duan2018deep,zheng2019hardness} to simultaneously decrease and increase intra-class and inter-class variances. Our algorithm falls in the first category where we design a novel, yet effective 
loss function that aims to decrease the overlap between the pairwise similarity distributions of the positive and the negative pairs for every class within the mini-batch.

\subsubsection*{Proposed Approach}
We pass the input images $\Mat{X}_i$ through the DNN to obtain their embedding points $\Mat{z}_i$. 
For a particular class label $k \in \mathcal{Y}$, we obtain two different sets of features embeddings 
$\left [ \Mat{Z} \right ]^+_k$ and $\left [ \Mat{Z} \right ]^-_k$ as shown below:
\begin{equation}
\label{eqn:grouping}
  \left[ \Mat{Z} \right]^+_k = \left\{ \Vec{z}_i \mid \Vec{y}_i=k \right \} ~~~~~~~~~~
    \left[ \Mat{Z} \right]^-_k = \left\{ \Vec{z}_i \mid \forall~\Vec{y}_i~~\&~~\Vec{y}_i \neq k \right \}
\end{equation}

\noindent$\left[ \Mat{Z} \right]^+_k$ and $\left[ \Mat{Z} \right]^-_k$ denote the embeddings that are class-specific and class-distinct for a particular class $k$. The overall $\mathrm{DCDL}$ loss is given as follows:
\begin{equation}
\label{eqn:train_loss}
\centering
    \mathrm{L}^{\varphi}  = - \sum_{k=1}^{\textrm{K}} \mathrm{L}_k~~~~~~~~~\textrm{where}~~~~ \mathrm{L}_k = \varphi \big(\left[ \Mat{Z} \right]^+_k, \left[ \Mat{Z} \right]^-_k \big)~~~,
\end{equation} where $\varphi$ denotes the probability discrepancy calculated using $\mathrm{MMD}$ with Laplacian or Gaussian kernels (as shown in Eqn.~(\textcolor{red}{16}) of the supplementary) 
or Wasserstein Distance (as shown in Eqn.~(\textcolor{red}{11}) of the supplementary).

An overall schematic of $\textrm{DCDL}$ (\ie $\mathrm{L}^{\varphi}$) is shown in Fig.~\ref{fig:ours_DCDL}, where DNN denotes a Deep Neural Network with learnable parameters $\Theta$. As an illustration, we have shown only $3$ classes in the input images and each have been assigned a different color, \ie \textcolor{red}{red}, \textcolor{blue}{blue} and \textcolor{ao}{green} respectively. \textbf{Black Double} arrows represent the main notion of $\mathrm{DCDL}$; \ie to enforce and increase the distance between the probability distributions of the class-wise embeddings for every single class (Eqn.~\eqref{eqn:grouping}).

\noindent \textbf{Compact Class-wise Distributions:} 
$\mathrm{L}^{\varphi}$ enforces a disparity constraint between the class-wise distributions, thereby ensuring that these class-wise distributions are well spread in the embedding space. However, it is to be noted that the class-wise distributions are non-parametric and are not explicitly modeled by $\mathrm{L}^{\varphi}$. Hence it becomes difficult to make such class-wise distributions dense and compact. Thus along with $\mathrm{L}^{\varphi}$, we have incorporated several well-studied \emph{local} loss functions (a) Triplet loss $\mathrm{L_{Trip}}$~\citep{Schroff_CVPR_2015}, (b) NPairs loss $\mathrm{L_{NP}}$~\citep{Sohn_NIPS_2016}, (c) Angular loss $\mathrm{L_{Ang}}$~\citep{wang2017deep} and (d) $\mathrm{L_{Ang\_NP}}$~\citep{wang2017deep}; to reduce the variances of each of the class-wise distribution so as to obtain tightly knit and well separated class-wise clusters. The various \emph{local} loss functions used are defined below:

\begin{equation}
\footnotesize
    \begin{aligned}
         \mathrm{L_{Trip}} &= \hspace*{-0.5ex} \frac{1}{|P|} \hspace*{-0.5ex} \sum_{m=1}^{|P|} \hspace*{-0.5ex} 
        	\Big[ \big \| \Vec{Z}_m^a \hspace*{-0.5ex} - \hspace*{-0.5ex} \Vec{Z}_m^+ \big \|^2  \hspace*{-1.5ex} - \hspace*{-0.5ex}
        	\big \| \Vec{Z}_m^a \hspace*{-0.5ex} - \hspace*{-0.5ex} \Vec{Z}_m^- \big \|^2 
        	\hspace*{-1.0ex} + \hspace*{-0.5ex} \tau \Big]_+ \hspace*{-0.5ex} ~~, 
        ~~~~\mathrm{L_{NP}} = \frac{1}{N}\!\! \sum_{\Vec{Z}^a \in \mathrm{B}} \Big\{ \mathrm{log}~ \Big[ 1+\!\!\!\!\!\!\sum_{\substack{\Vec{Z}^n \in \mathrm{B} \\ \Vec{y}^n \neq \Vec{y}^a, \Vec{y}^p}}\!\!\!\!\!\!\! \mathrm{exp}  ( \Vec{Z}^{a^\top}\Vec{Z}^{-} - \Vec{Z}^{a^\top}\Vec{Z}^{+} ) \Big]  \Big\} ~~, \\
        \mathrm{L_{Ang}} &= \frac{1}{N}\!\! \sum_{\Vec{Z}^a \in \mathrm{B}}\!\!\! \Big\{\! \mathrm{log}~\!\! \Big[1~+\!\!\!\!\sum_{\substack{\Vec{Z}^{-} \in \mathrm{B} \\ \Vec{y}^{-} \neq \Vec{y}^{a}, \Vec{y}^{+}}}\!\!\!\!\!\!\! \mathrm{exp}\big( \!~4~\mathrm{tan}^2\alpha(\Vec{Z}_a+\Vec{Z}^{+})^\top\Vec{Z}^{-}
        - \!~2~(1+\mathrm{tan}^2\alpha)\Vec{Z}^{a^\top}\!\Vec{Z}^{+} \big) \Big] \Big\} ~~,
    \end{aligned}
\end{equation}
where (\RNum{1}) $[y]_+ = \max(0,y)$ is the hinge loss, (\RNum{2}) $\tau > 0$ is a user-specified margin, (\RNum{3}) $|P|$ represents the number of triplets of the form $(\Vec{Z}^{a}, \Vec{Z}^{+}, \Vec{Z}^{-})$ such that $\Vec{y}^n\!\neq\!\Vec{y}^a$ and $\Vec{y}^a\!=\!\Vec{y}^p$, (\RNum{4}) $\alpha$ is a predefined parameter constraining the angle formed at $\Vec{Z}^{-}$ by the triplet $(\Vec{Z}^{a}, \Vec{Z}^{+}, \Vec{Z}^{-})$ and (\RNum{5}) $\mathrm{B}$ denotes the batch of the samples. Additionally, as per the suggestion in~\citep{wang2017deep}, we also train our model with a combination of $\mathrm{L_{NP}}$ and $\mathrm{L_{Ang}}$ as shown below:
\begin{equation}
\mathrm{L_{Ang\_NP}} = \mathrm{L_{NP}} + \lambda_{Ang}~\mathrm{L_{Ang}}~~~,
\end{equation}
where $\lambda_{Ang}$ is set to $2$ in our experiments. 

\par\noindent\textbf{Training Loss:}
As observed, the aforementioned local loss functions enforce \emph{local} constraints in their respective formulation and fail to enforce the 
dissimilarity constraints in learning a discriminative embedding space. Therefore, we augment these loss functions with $\mathrm{L}^{\varphi}$ 
so as to holistically integrate these constraints in terms of a class-wise discrepancy measure within the mini-batch $\mathrm{B}$. The final training loss is defined as follows:
\begin{equation}
\label{eqn:train_loss_final}
    \mathrm{L_{Train}} = \mathrm{L_{Local}} + \lambda~\mathrm{L}^{\varphi} 
\end{equation}
where $\mathrm{L_{Local}}$ is either $\mathrm{L_{Trip}}, ~\mathrm{L_{NP}},~\mathrm{L_{Ang}}$ or $\mathrm{L_{Ang\_NP}}$ corresponding to Triplet, NPairs, Angular and the combination of NPairs and Angular loss respectively\footnote{Similar to $\mathrm{L_{Train}}$, \citep{lin2018deep} also make use of a combination of various metric loss functions along with KL divergence loss to learn well separated but compact clusters}.

%% file: SOUMAVA-FILES/paper_related.tex
\section{Related Work}
\label{sec:eccv2020_related}
In this section we provide a brief overview of the several baseline Deep Metric Learning algorithms which have been developed in the recent past.%

\par\noindent\textbf{Structure based methods:} The seminal work in DML was undertaken by Schroff \etal~\citep{Schroff_CVPR_2015}; where they proposed a \emph{semi-hard} triplet mining algorithm which guarantees that for a given \textit{anchor} $(\Vec{z}^{a})$, its closest \textit{negative} $(\Vec{z}^{-})$ is further away from the \textit{positive} $(\Vec{z}^{+})$. Similar constraints in the angle formed at $\Vec{z}^{-}$ within a triplet in the embedding space was proposed by Wang~\etal~\citep{wang2017deep}. Hierarchical Triplet loss proposed by Weifeng~\etal~\citep{ge2018deep} constructed a class-level tree and dynamically learnt a margin such that the \emph{semi-hard} triplet mining criterion is fulfilled. Song~\etal~\citep{Song_CVPR_2017} proposed a structured prediction loss to enforce global geometrical constraints in the embedding space to prevent outlier clusters. Roy~\etal~\citep{roy2019siamese} imposed rotation-invariant orthogonality constraints in the embedding space to truncate the search space for learning an efficient metric. Law~\etal~\citep{Law_ICML_2017} exploited spectral clustering concepts to obtain tight but well separated clusters in the embedding space. Sanakoyeu~\etal~\citep{sanakoyeu2019divide} splits the embedding space into $\textrm{P}$ consecutive parts; and learnt a local discriminative loss over each of the individual parts with the aim to learn distinct and disjoint features in each of them. SoftTriple loss~\citep{qian2019softtriple} additionally models and learns multiple clusters per class as an additional layer and is trained with a smoothened version of the triplet loss.

\par\noindent\textbf{Sampling based methods:} The objective of these algorithms is to learn efficient sampling strategies with the aim to mine important and informative samples. Lifted-Structure proposed by Song~\etal~\citep{Song_CVPR_2016} constructs a mini-batch of samples by subsequent addition of importance-sampled hard negatives. Sohn~\citep{Sohn_NIPS_2016} proposed a multi-class \emph{NPairs} loss by utilizing all the negatives for every anchor point $\Vec{z}^a$. They also proposed an efficient batch construction scheme that uses all the available negative pairs in order to form the informative triplets. Wu~\etal~\citep{Wu_ICCV_2017} clearly demonstrated the importance of a steady sampling function to mine informative negative samples to reduce noisy back-propagated gradients.

\par\noindent\textbf{Statistical based methods} aim to model (parametric or non-parametric) the statistical distributions in the embedding space. Histogram loss~\citep{Ustinova_NIPS_2016} enforces an overlap constraint minimizing the overlap between the histograms of the cosine similarities for the positive and the negative pairs respectively. Kumar~\etal~\citep{kumar2016learning} formulates the pairwise distances between the matching and non-matching pairs as two different Gaussians; and proposes the Distribution loss to reduce the overlap between them. Roth~\etal~\citep{roth2019mic} learnt an auxiliary encoder to reduce the intra-class variances while separating the intra-class means of the class-wise embedding points.

\par\noindent\textbf{Generative modeling based methods:} Deep Adversarial Metric Learning~\citep{duan2018deep} learns a Generative Adversarial Network~\citep{goodfellow2014generative} to generate hard negatives to enable efficient learning of the discriminative embedding space. Lin~\etal~\citep{lin2018deep} proposed Deep Variational Metric Learning that makes use of variational inference to generate robust discriminative samples to reduce the intra-class variances. Zheng~\etal~\citep{zheng2019hardness} proposed to control the hardness level of the generated negatives by linear interpolation over the entire embedding space, thereby resulting in well spread out clusters of the embedding samples.

\begin{remark}
It is to be noted that none of the above mentioned algorithms comprehensively model class-wise similarity relationships while learning the embedding space. Thus, it seems that the embedding points are neither tightly knit nor well spread over the entire embedding space. Unlike these algorithms, $\mathrm{DCDL}$ learns class-wise distributions belonging to each and every distinct class and maximizes the separation between them to group the class-wise embeddings within certain specific regions of the embedding space while retaining their class-wise relationships.
\end{remark}


\begin{remark}
Statistical based methods learn a discriminative embedding by enforcing similarity (or dissimilarity) constraints on the probability distributions of all the positive and negative pairs within the dataset. However as before, these methods do not consider class-wise similarity/dissimilarity constraints between the positive and the negatives pairs for each individual class within the dataset. Moreover, they enforce a weak discrepancy measure to reduce the overlap between the probability distributions and hence are not able to decrease (increase) intra-class (inter-class) variances within the dataset, thus obtaining a sub-optimal embedding space. On the other hand, $\mathrm{DCDL}$ \D{considers the class-wise samples within each mini-batch to calculate such class-wise probability distributions.} Thus, apart from maintaining class-wise relationships, $\mathrm{DCDL}$ also employs the well-known Optimal Transport probability diversity measures as a discrepancy function to reduce the overlap between these distributions~\D{in order to increase the inter-class variances within the embedding space. Moreover, we also add a local metric learning loss function to $\mathrm{DCDL}$ in order to reduce the intra-class variances within each of the probability distributions (Refer to Eqn.~\eqref{eqn:train_loss_final} in Section~\textsection~\ref{sec:eccv2020_DCDL})} Further, the statistical methods mentioned above are highly sensitive to their respective parameters (such as number of histograms bins in~\citep{Ustinova_NIPS_2016}, dimensionality of the encoder in~\citep{roth2019mic} \textit{etc.}) which are used to empirically estimate the probability distributions; while Optimal Transport is less sensitive to its corresponding hyper-parameters~\citep{cuturi2013sinkhorn}.
\end{remark}

%% file: SOUMAVA-FILES/paper_simple_exps.tex
\section{Experiments}
\label{sec:eccv2020_expts}
\subsection*{Image Classification Datasets:}
We evaluate our proposed loss on the following well-studied image classification datasets:
\begin{enumerate}
    \item Cifar-$10$~\citep{krizhevsky2009learning} ($\textrm{C}$-$10$) consists of $32 \times 32 \times 3$ images from $10$ different categories. The training and the test set consist of $50,000$ and $10,000$ images respectively. %
    \item Cifar-$100$~\citep{krizhevsky2009learning} ($\textrm{C}$-$100$)  consists of $32 \times 32 \times 3$ images from $100$ different \emph{fine} categories, which can be further grouped into $20$ super categories. We have considered the $20$ super categories in our experiments. Similar to Cifar-$10$~\citep{krizhevsky2009learning}, the training and the test set consist of $50,000$ and $10,000$ images respectively. 
    \item STL-$10$~\citep{coates2011analysis}  ($\textrm{S}$-$10$) is an image recognition dataset that consists of $96 \times 96 \times 3$ images from $10$ different classes, each comprising $1,300$ examples. The training set consist of $5,000$ images, while the remaining $8,000$ images constitute the test set. It also consists of $100,000$ unlabeled images of the same resolution extracted from a similar but a wider distribution of images. However, we don't use these unlabeled images, neither during the training nor during the test phase in our proposed algorithm. 
\end{enumerate}
\vspace{-0.5cm}
\subsection*{Implementation Details}
We implemented $\mathrm{DCDL}$ in PyTorch~\citep{paszke2017automatic}. We used VGG-9~\citep{li2018visualizing} as the backbone network with a fully-connected, embedding layer at the end. We randomly initialize~($\mathrm{U}\left [0,1\right])$) all the layers in the network. The dimension of the embedding layer is fixed to $64$. A dropout layer with a dropout rate of $0.1$ is added before the embedding layer. When $\mathrm{L_{Local}}\!\!=\!\!\mathrm{L_{Trip}}$, we ensure that there are at least $10$ data samples from every class; thereby resulting in a mini-batch of $100$ for the Cifar-$10$ and STL-$10$ datasets, and $200$ for the Cifar-$100$ dataset respectively. Likewise, when $\mathrm{L_{Local}}\!=\!\left \{\mathrm{L_{NP}},\mathrm{L_{Ang}}, \mathrm{L_{Ang\_NP}} \right \}$, we guarantee that there are at-least $2$ data samples from every class within a mini-batch. This leads to a mini-batch of size $20$ for the Cifar-$10$ and STL-$10$ datasets, and $40$ for the Cifar-$100$ dataset respectively for these loss functions. We have not used any data augmentation techniques for the train and test sets across all the datasets\footnote{The training and test images of STL-$10$ are resized to $32 \times 32 \times 3$.}. In our empirical evaluations with $\mathrm{L_{NP}}$, we have observed that the Stochastic Gradient Descent (SGD) optimizer with Momentum attained the best accuracy on the test set in the Cifar-$10$ dataset; whereas for the others we have used the Adam~\citep{kingma2014adam} optimizer to fine-tune the parameters of the network. The initial learning rate for the Adam and the SGD optimizer is set to $5\!\times\!10^{-4}$ and $1\!\times\!10^{-2}$ respectively. We decay the learning rate by a factor of $0.1$ after every $50$ epochs. We compare and evaluate our proposed algorithm (\ie $\mathrm{DCDL}$) against several baseline algorithms, namely (\RNum{1}) Trip~\citep{Schroff_CVPR_2015}, (\RNum{2}) NPairs (NP)~\citep{Sohn_NIPS_2016}, (\RNum{3}) Angular (Ang)~\citep{wang2017deep} and (\RNum{4}) Angular-Npairs (Ang\_NP)~\citep{wang2017deep}. In all of our experiments, we train our models for $100$ epochs with $\mathrm{L_{Train}}$ as defined in Eqn.~\eqref{eqn:train_loss}. Further, we evaluate the discriminative ability of an embedding space by learning a linear classifier on the embeddings of the training set and report the classification accuracy on the test set. We set the value of $\epsilon$ to $2.5\!\times\!10^{-3}$~\citep{feydy2019interpolating} for experiments with $\mathrm{L}^{\mathcal{W}}$(Refer to Eqn.(~\textcolor{red}{11}) of the supplementary).
Further, we set the value of \textbf{(a)} $\tau$ in $\mathrm{L_{Trip}}$ to $0.5$, \textbf{(b)} $\alpha$ in $\mathrm{L_{Ang}}$ and $\mathrm{L_{Ang\_NP}}$ to $\ang{30}$ and $\ang{45}$ respectively. We report the best results (\ie the classification accuracy in \% on the test set) obtained for all the hyper-parameter settings after training the models for $100$ epochs for all the datasets across different experimental settings.

\input{SOUMAVA-FILES/paper_figs}

\subsection*{Results}
\label{sec:eccv2020_results}
\textbf{Notations:} While reporting the quantitative results, we have used the following notation to avoid any confusion. We represent the calculation of $\mathrm{L}^{\varphi}$ with \underline{L}aplacian and \underline{G}aussian kernels for $\mathrm{MMD}$ as $\mathrm{L}^{\mathrm{L}}$ and $\mathrm{L}^{G}$ respectively; with the value of $\sigma$ fixed to $0.05$ across all the experiments~(refer to Eqn.~(\textcolor{red}{16}) of the supplementary)
~\citep{Han_2020_WACV}. Similarly, $\mathrm{L}^{\mathcal{W}}$ denotes the calculation of $\mathrm{L}^{\varphi}$ with $\mathcal{W}_\epsilon^{p}(\mathcal{U}, \mathcal{V})$ such that $p\!=\!2$ and $\Mat{D}^{p}(\Vec{u}, \Vec{v})\!=\! \frac{1}{2} \left \|\Vec{u}-\Vec{v}\right \|_2^2$)~(refer to Eqn.~(\textcolor{red}{11}) of the supplementary). $\sigma$ and $\epsilon$ are set to $0.05$, $2.5\times10^{-3}$ respectively. The dimension of the embedding is set to $64$. The value of $\lambda$ is set to $0.2$ and $0.5$ for all the experiments with $\mathrm{L}^{\mathrm{L}}$ and $\mathrm{L}^{\mathcal{W}}$ respectively. We also sample $10$ samples per class for every dataset. A detailed analysis of selecting these value is present at Section~\textsection~\textcolor{red}{2} of the supplementary material. We report the test accuracy for experiments with \textbf{noisy} labels
; where we artificially corrupt the ground truth labels of the training set across all the datasets, while the test labels are unaltered. The transition of labels is parameterized by $\delta \in \left [0, 1\right]$ such that the probability of the true and the noisy class labels are $\delta$ and $1-\delta$ respectively. Noisy labels can be further categorized as follows: 
\begin{itemize}
    \item \textbf{Symmetric (Sym) Noise:} Similar to~\citep{li2019learning}, in this experimental setup we corrupt the ground truth train labels with a random one-hot vector with a probability of $1-\delta$. An exemplar noise transition matrix for the symmetric noise is shown in Fig.~\ref{fig:noise_transition}~\textbf{(a)}. We evaluate $\mathrm{DCDL}$ for two values of $\delta$, \ie, $\{0.1, 0.3\}$ for each of the three datasets.
    
    \item  \textbf{Asymmetric (Asym) Noise:} Asymmetric noise takes into account realistic real life mistakes, thus confusing similar classes within the dataset. Similar to the asymmetric noise protocol of~\citep{li2019learning}, we also create the following class transition pairs for the Cifar-$10$ and STL-$10$ datasets\footnote{We do not evaluate on the Cifar-$100$ dataset for the Asymmetric Noise setting.}; (\RNum{1}) TRUCK $\rightarrow$ AUTOMOBILE, (\RNum{2}) BIRD $\rightarrow$ AIRPLANE, (\RNum{3}) DEER $\rightarrow$ HORSE and (\RNum{4}) CAT $\leftrightarrow$ DOG. An exemplar noise transition matrix for the asymmetric noise is shown in Fig.~\ref{fig:noise_transition}~\textbf{(b)}. The value of $\delta$ is set to $\{0.1, 0.2\}$ in our experiments for this setting.
\end{itemize}

\begin{table}[t]
\centering
    \caption{\small Experimental results demonstrating the importance of incorporating $\mathrm{L}^{\mathrm{L}}$ and $\mathrm{L}^{\mathcal{W}}$ in Eqn.~\eqref{eqn:train_loss_final} across the three datasets for different values of $\delta$ in the \textbf{symmetric} and \textbf{asymmetric} noisy label setting.}
    \label{tab:noisy_both}
\vspace{0.2cm}
\scalebox{0.9}{
\begin{tabular}{|c|ccc|ccc|cc|cc|}
\toprule
 & \multicolumn{6}{c|}{Symmetric Noise} & \multicolumn{4}{c|}{Asymmetric Noise} \\
 \midrule
Setting & \multicolumn{3}{c|}{0.1} & \multicolumn{3}{c|}{0.3} & \multicolumn{2}{c|}{0.1} & \multicolumn{2}{c|}{0.2} \\
\midrule
Dataset & C-10 & C-100 & S-10 & C-10 & C-100 & S-10 & C-10 & S-10 & C-10 & S-10 \\
\midrule
$\mathrm{L_{Trip}}$ & 80.4 & 58.9 & 48.4 & 71.1 & 33.8 & 39.8 & 82.5 & 55.2 & 78.4 & 54.6 \\
$\mathrm{L_{Trip}}$ + $\mathrm{L}^{\mathrm{L}}$ & 82.2 & 60.9 & 51.8 & 72.7 & 34.7 & 41.2 & 84.1 & 57.5 & 79.7 & 55.8 \\
$\mathrm{L_{Trip}}$ + $\mathrm{L}^{\mathcal{W}}$ & 82.8 & 61.9 & 52.5 & 74.3 & 36.7 & 43.6 & 84.9 & 59.0 & 81.7 & 57.9 \\
\midrule
$\mathrm{L_{NP}}$ & 65.5 & 36.9 & 39.8 & 51.1 & 23.3 & 32.9 & 64.2 & 43.7 & 53.9 & 42.1 \\
$\mathrm{L_{NP}}$ + $\mathrm{L}^{\mathrm{L}}$ & 65.4 & 37.5 & 44.0 & 50.9 & 26.2 & 35.2 & 65.9 & 46.4 & 55.2 & 45.5 \\
$\mathrm{L_{NP}}$ + $\mathrm{L}^{\mathcal{W}}$ & 67.1 & 38.1 & 48.4 & 51.4 & 27.4 & 43.2 & 67.2 & 49.8 & 56.9 & 47.3 \\
\midrule
$\mathrm{L_{Ang}}$ & 80.8 & 49.9 & 46.8 & 63.2 & 25.1 & 35.2 & 82.0 & 53.5 & 78.6 & 50.6 \\
$\mathrm{L_{Ang}}$ + $\mathrm{L}^{\mathrm{L}}$ & 81.4 & 51.1 & 48.1 & 64.8 & 26.3 & 37.8 & 83.3 & 55.1 & 79.7 & 51.3 \\
$\mathrm{L_{Ang}}$ + $\mathrm{L}^{\mathcal{W}}$ & 82.1 & 53.7 & 49.3 & 66.3 & 28.6 & 40.8 & 85.1 & 56.2 & 82.0 & 53.6 \\
\midrule
$\mathrm{L_{Ang\_NP}}$ & 80.7 & 49.5 & 46.3 & 64.2 & 24.6 & 35.7 & 81.7 & 52.1 & 78.2 & 50.9 \\
$\mathrm{L_{Ang\_NP}}$ + $\mathrm{L}^{\mathrm{L}}$ & 81.6 & 51.2 & 47.2 & 65.5 & 25.8 & 37.3 & 82.5 & 52.7 & 79.6 & 50.3 \\
$\mathrm{L_{Ang\_NP}}$ + $\mathrm{L}^{\mathcal{W}}$ & 82.1 & 52.7 & 48.7 & 67.1 & 28.1 & 38.5 & 83.2 & 54.6 & 81.7 & 53.9 \\
\bottomrule
\end{tabular}
}
\vspace{-0.5cm}
\end{table}


Table~\ref{tab:noisy_both} shows the results reported on the three datasets for various configurations of noise and $\mathrm{L_{Local}}$ in the symmetric and asymmetric noisy labels settings respectively. We can easily observe that barring a few configuration settings; the accuracy obtained with symmetric noise is lower than the one obtained via asymmetric noise corruption. It is not surprising as the transitional categories for asymmetric noise are visually quite similar to each other, thus so are the features extracted and learnt by our network; thereby leading to improved results over symmetric-noise settings. Moreover, it is also observed that adding $\mathrm{L}^{\mathrm{L}}$ and $\mathrm{L}^{\mathcal{W}}$ to $\mathrm{L_{Local}}$ outperform its corresponding conventional local loss functions (\ie $\mathrm{L_{Local}}$) by a considerable margin in terms of classification accuracy, thereby comprehensively demonstrating the need of enforcing such class-wise discrepancy constraints in learning a superior embedding in the noisy labels settings. 


%% file: SOUMAVA-FILES/paper_figs.tex
\begin{figure*}[!t]
\begin{minipage}[h]{0.45\linewidth}
        \centering
        \begin{tabular}{cccccc|}
         & \multicolumn{1}{c}{0} & \multicolumn{1}{c}{1} & \multicolumn{1}{c}{2} & \multicolumn{1}{c}{3} & \multicolumn{1}{c}{4} \\ \cline{2-6} \cline{2-6} 
        0 & \multicolumn{1}{|c|}{\cellcolor{blue!25}$\delta$} & \cellcolor{Gray} & \cellcolor{Gray} & \cellcolor{Gray} & \cellcolor{Gray} \\ \cline{2-3}
        1 & \multicolumn{1}{|c|}{\cellcolor{Gray}} & \multicolumn{1}{c|}{\cellcolor{blue!25}$\delta$} & \multicolumn{3}{c|}{\cellcolor{Gray}} \\  \cline{3-4}
        2 & \multicolumn{2}{|c}{\cellcolor{Gray}$1-\delta$} & \multicolumn{1}{|c|}{\cellcolor{blue!25}$\delta$} & \multicolumn{2}{c|}{\cellcolor{Gray}$1-\delta$} \\  \cline{4-5}
        3 & \multicolumn{3}{|c}{\cellcolor{Gray}} & \multicolumn{1}{|c|}{\cellcolor{blue!25}$\delta$} & \cellcolor{Gray} \\ \cline{5-6} 
        4 & \multicolumn{4}{|c}{\cellcolor{Gray}} & \multicolumn{1}{|c|}{\cellcolor{blue!25}$\delta$} \\ \cline{2-6} 
        \end{tabular}
        \caption*{\hspace{0.2cm}(a)}
        
        \begin{tabular}{cccccc}
         & \multicolumn{1}{c}{0} & \multicolumn{1}{c}{1} & \multicolumn{1}{c}{2} & \multicolumn{1}{c}{3} & \multicolumn{1}{c}{4} \\
         \cline{2-2}  \cline{6-6}
        \multicolumn{1}{c|}{0} & \multicolumn{1}{|c|}{\cellcolor{blue!25}$\delta$} & \cellcolor{Gray1} & \cellcolor{Gray1} & \cellcolor{Gray1} & \multicolumn{1}{|c|}{\cellcolor{Gray}1-$\delta$} \\ 
        \cline{2-4} \cline{6-6}
        1 & \cellcolor{Gray1} & \multicolumn{1}{|c|}{\cellcolor{blue!25}$\delta$} & \multicolumn{1}{c|}{\cellcolor{Gray}1-$\delta$} & \cellcolor{Gray1} & \cellcolor{Gray1} \\ \cline{3-5} \cline{3-5}\cline{3-5} 
        2 &\cellcolor{Gray1} & \multicolumn{1}{c|}{\cellcolor{Gray1}} & \multicolumn{1}{c|}{\cellcolor{blue!25}$\delta$} & \multicolumn{1}{c|}{\cellcolor{Gray}1-$\delta$} & \cellcolor{Gray1} \\ \cline{4-6} \cline{4-6} 
        3 & \cellcolor{Gray1} & \cellcolor{Gray1} & \multicolumn{1}{c|}{\cellcolor{Gray1}} & \multicolumn{1}{c|}{\cellcolor{blue!25}$\delta$} & \multicolumn{1}{c|}{\cellcolor{Gray}1-$\delta$} \\ \cline{2-2} \cline{5-6} \cline{2-2} \cline{5-6}
        4 & \multicolumn{1}{|c|}{\cellcolor{Gray}1-$\delta$} & \cellcolor{Gray1} & \cellcolor{Gray1} & \multicolumn{1}{c|}{\cellcolor{Gray1}} & \multicolumn{1}{c|}{\cellcolor{blue!25}$\delta$} \\ \cline{2-2} \cline{6-6} 
        \end{tabular}
        \caption*{\hspace{0.3cm}(b)}
\end{minipage}
\hspace{0.5cm}
\begin{minipage}[h]{0.4\linewidth}
\centering
    \includegraphics[width=1\textwidth,keepaspectratio]{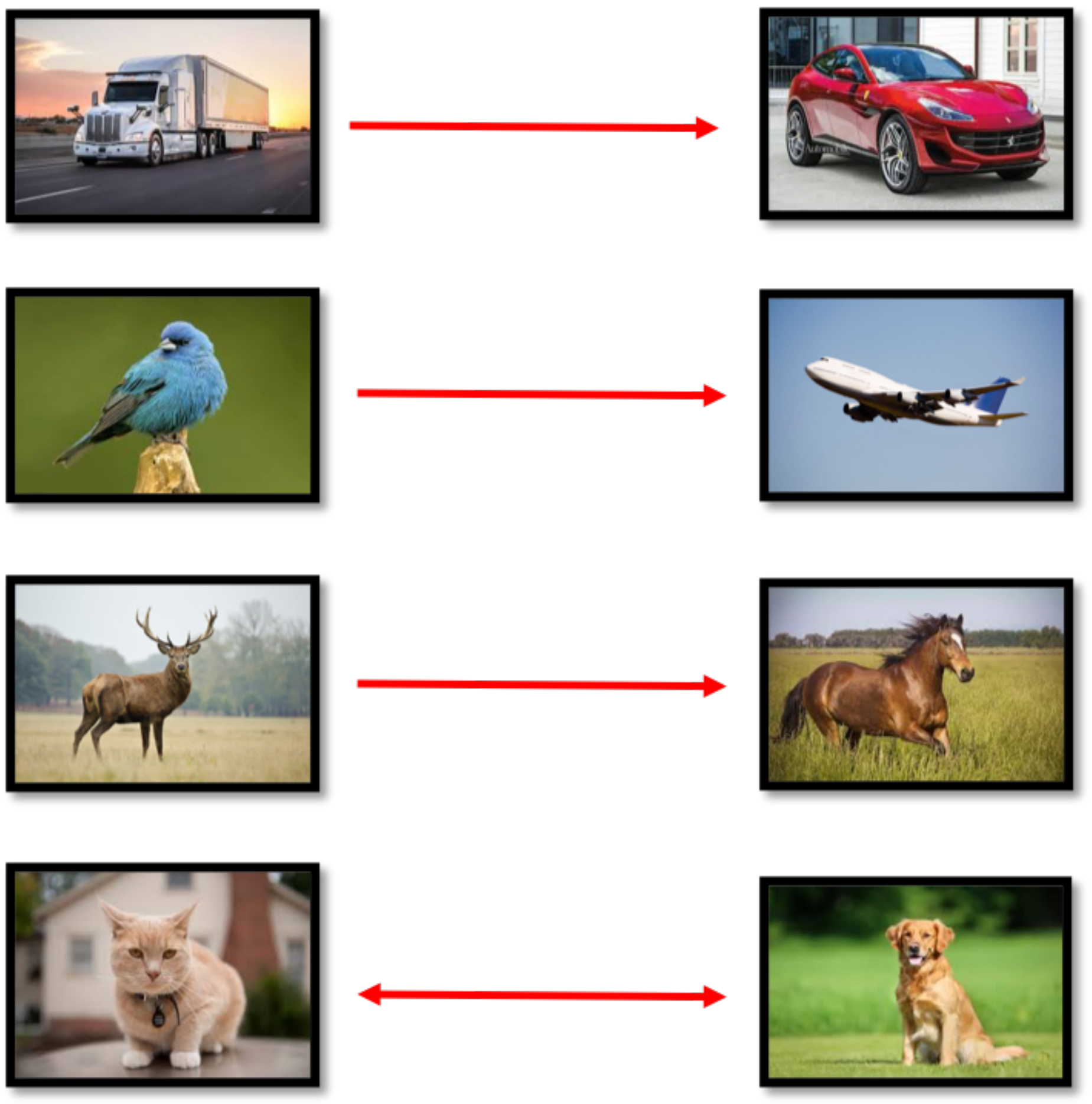}
    \caption*{\hspace{0.1cm}(c)}
\end{minipage}
\vspace{0.2cm}
\caption{\small Exemplar Noise Transition matrix for $5$ classes for (a) \textbf{Symmetric} and (b) \textbf{Asymmetric} Noise. $\delta$ denotes the percentage of noise for both settings. (c) Transition class pairs for Asymmetric noise.}
\label{fig:noise_transition}
\vspace{-0.5cm}
\end{figure*}

%% file: SOUMAVA-FILES/paper_conclusion.tex
\vspace{-0.5cm}
\section{Conclusion and Future Work}
\label{sec:eccv2020_conclusion}
\vspace{-0.2cm}
Here, we proposed a novel Deep Class-wise Discrepancy Loss ($\mathrm{DCDL}$) and addressed the problem of learning an efficient and discriminative embedding space. To this end, we have incorporated and increased \emph{class-wise} dissimilarity constraints so as to decrease the intra-class variances, and increase the inter-class variances thereby leading to the formation of tight but well separated clusters of embedded points. Towards achieving this objective, we employed Maximum Mean Discrepancy and Wasserstein Distance, with roots in Optimal Transport, in order to separate the probability distributions between the embeddings belonging to each and every class. We have evaluated our proposed methodology across \textbf{noisy} label settings over three well known datasets. Our empirical evaluations support and substantiate the need of enforcing such global dissimilarity constraints in learning a superior embedding over the baseline algorithms. A similar trend was also observed while training with an additional partially-global cross entropy loss; thereby further reinforcing the necessities of utilizing $\mathrm{DCDL}$ in learning a tightly knit embedding space. We have also demonstrated superior performance of $\textrm{DCDL}$ in the two fine-grained image classification datasets for both clean and noisy labels settings. Our empirical evaluations showed that global class-wise divergences are needed to learn an efficient and effective embedding space. In the future, we aim to design a mutually integrated loss function to further improve the learning ability of $\textrm{DCDL}$.

%% file: SOUMAVA-FILES/paper_prelims.tex
\section{Preliminaries}
\label{sec:prelim}
Estimating the parameters (\ie~$\theta \in \Theta$) of probability density functions chosen from a family of parametric distributions $\textrm{p}_\Theta$ in order to fit the observed data in the best possible and meaningful way is the fundamental approach in several machine learning algorithms. One such popular algorithm is Maximum Likelihood Estimation (MLE); which aims to obtain the optimal parameters \ie $\theta^{*}$ that maximizes the likelihood of the observed data as shown below:
\begin{equation}
    \theta^* \vcentcolon= \underset{\theta \in \Theta}{\textrm{max}}~\frac{1}{N}~\sum_{i=1}^{N}~\ln ~\textrm{p}_\theta(\Mat{X}_i) ~~,
\end{equation}
where $\left \{ \Mat{X}_i \right \}_{i=1}^{N}$ denote the $N$ observed data samples. Though highly popular and successful, MLE fails in those instances where the density of the desired distribution is singular on the observation space. One such example being generative modeling algorithms such as Generative Adversarial Networks (GANs)~\citep{goodfellow2014generative}; Variational Auto-Encoders (VAEs)~\citep{kingma2013auto} which rely on a sampling mechanism to map a low dimensional random vector to a high dimensional space. Nonetheless, GANs attempt to overcome this limitation by utilizing classification accuracy as a proxy for similarity between two parameterized distributions; whereas VAEs use Kullback–Leibler divergence (KL)~\citep{kullback1951information} to reduce the distance between two parametric distributions. However, such practices fail to consider and respect the underlying geometry of the observation space. Therefore, GANs suffer from mode collapse and vanishing gradients~\citep{srivastava2017veegan}; whereas VAEs fail to provide any meaningful representation due to mismatch between two non-overlapping distributions.

\subsection*{Optimal Transport}

Optimal Transport (OT)~~\citep{genevay2018learning}, endowed with its own distance metric, provides a natural and an effective framework for measuring the distance between two probability distributions. Specifically, OT measures the amount of distance traveled in transporting the mass from one distribution so as to look similar to the other distribution. OT considers the underlying geometry of the observation space while calculating the distance; thereby making it better suited to capture complex similarity relations between objects in the observation space when compared to other distance metrics (such as the Euclidean distance).

Given two probability distributions $\mathcal{U}$ and $\mathcal{V}$ defined over the metric space $\mathcal{M}$ with probability measures $\mu$ and $\nu$ respectively, OT aims to find an \emph{optimal transport plan} $\omega^* \in \Omega(\mu, \nu)$ 
in order to reduce some notion of a cost function while transferring mass from $\mathcal{U}$ to $\mathcal{V}$. The formulation is shown below: 
\begin{equation}
    \omega^* \vcentcolon= \underset{\omega \in \Omega(\mu, \nu)}{\textrm{inf}} \left \{ \int_{\mathcal{M} \times \mathcal{M}} \mathrm{c}(\Vec{u},\Vec{v})~d\omega(\Vec{u}, \Vec{v}) \right \} ~~~,
\end{equation}
where $\mathrm{c}(\Vec{u},\Vec{v})$ represents the cost of transferring a unit mass from the support point $\Vec{u}$ to $\Vec{v}$ 
over the metric space $\mathcal{M}$. $\Omega(\mu, \nu)$ represents the overall set of all possible transport plans in order to distribute mass from $\mathcal{U}$ to $\mathcal{V}$.

\subsection*{$p$-Wasserstein Distance}
Given $p \geq 1$ and the cost function $\mathrm{c}(\Vec{u}, \Vec{v}) \vcentcolon= \textrm{d}(\Vec{u}, \Vec{v})^{p}$ (where $\textrm{d}$ is a distance metric~\citep{weinberger2009distance}); we obtain the $p$-Wasserstein Distance which is defined below:
\begin{equation}
\label{eqn:p_wasserstein}
\mathcal{W}^{p}(\mathcal{U}, \mathcal{V}) = \Big(~~ \underset{\omega~\in \Omega(\mu, \nu)}{\textrm{min}} \int_{\mathcal{M} \times \mathcal{M}} \!\!\!\textrm{d}(\Vec{u}, \Vec{v})^{p}~d\omega(\Vec{u}, \Vec{v}) ~\Big)^{1/p}
\end{equation}

\noindent\textbf{Note:} Here we consider both $\mathcal{U}$ and $\mathcal{V}$ to be \textbf{discrete} distributions and embed the feature vectors as finite sets of support points, \ie $\{\Vec{u}_i\}_{i = 1}^n$ ,$\{\Vec{v}_j\}_{j = 1}^m \subset \mathbb{R}^{l}$ ($l$ is the dimensionality of the metric space $\mathcal{M}$); for both the distributions respectively. Therefore, we have 
\begin{equation}
\label{eqn:discretization}
\begin{aligned}
\mathcal{U} &= \sum_{i=1}^{n} \Vec{w}_i \delta_{\Vec{u}_i} ~~~~,~~~~ \mathcal{V} = \sum_{j=1}^{m} \Vec{\widetilde{w}}_j \delta_{\Vec{v}_j}~~~, \\
\end{aligned}    
\end{equation}
such that the weights $\Vec{w}$ and $\Vec{\widetilde{w}}$ are constrained to be non-negative, and $\sum_{i=1}^{n}\!\Vec{w}_i\!=\!1$ and $\sum_{j=1}^{m}\!\Vec{\widetilde{w}}_j\!=\!1$. As a result of such discretization, the transport plan~$\Omega$ also becomes a discrete set $\mathrm{T}$ defined over the product of $\{\Vec{u}_i\}_{i = 1}^n$ and $\{\Vec{v}_j\}_{j = 1}^m$. Further, we also define $\Mat{D}^{p}\in\mathbb{R}_+^{n \times m}$ where $\Mat{D}^{p}_{ij}=\textrm{d}(\Vec{u}_i, \Vec{v}_j)^{p}$ (where $\textrm{d}$ is a global distance function). Therefore, for discrete distributions we obtain the following formulation from Eqn.~\eqref{eqn:p_wasserstein} as:
\begin{equation}
\label{eqn:discrete_ot}
\mathcal{W}^{p}(\mathcal{U}, \mathcal{V}) = 
\underset{\mathrm{T} \geq 0}{\textrm{min}}~~\textrm{Tr}(\Mat{D}^{p} ~\mathrm{T}^\top)
\end{equation}
where~$\mathrm{T}\Vec{1}\!=\!\Vec{w}$ and $\mathrm{T}^\top\Vec{1}\!=\!\Vec{\widetilde{w}}$. Eqn.~\eqref{eqn:discrete_ot} is computationally complex as the solution is a linear program. Thus, Cuturi in~\citep{cuturi2013sinkhorn} proposed an entropy regularized version of it such that the solution is more tractable, which is shown below:
\begin{equation}
\label{eqn:sinkhorn}
\mathcal{W}_\epsilon^{p}(\mathcal{U}, \mathcal{V})=\underset{\mathrm{T} \geq 0}{\min}~\textrm{Tr}(\Mat{D}^{p} ~\mathrm{T}^\top) +\epsilon~\textrm{Tr}\big(\mathrm{T}(\mathrm{ln}(\mathrm{T}) - \Vec{1}\Vec{1}^\top)^\top\big)
\end{equation}
where $\mathrm{T}\Vec{1}\!=\!\Vec{w}$, $\mathrm{T}^\top\Vec{1}\!=\!\Vec{\widetilde{w}}$, and $\epsilon$ is the regularization parameter. \D{Due to such entropy regularization, it is well known there exist $\Vec{\tilde{r}} \in \mathbb{R}_+^{n}$ and $\Vec{\tilde{c}} \in \mathbb{R}_+^{m}$ such that the optimal solution is $\textrm{T}^*\!=\!\Delta(\Vec{\tilde{r}})~\textrm{exp}(\frac{-\Mat{D}^{p}}{\epsilon})~\Delta(\Vec{\tilde{c}})$.} Therefore, instead of optimizing over all the possible outcomes of $\mathrm{T}$, one can optimize over $\Vec{r}$ and $\Vec{c}$ until convergence by alternatively projecting onto their respective marginals~\citep{sinkhorn1967diagonal} as follows:
\begin{equation}
\label{eqn:sinkhorn_iter}
    \Vec{r}^{t+1} \leftarrow \Vec{w} ./ \Mat{K} \Vec{c}^{t} ~~~~~, ~~~~~ \Vec{c}^{t+1} \leftarrow \Vec{\widetilde{w}} ./ \Mat{K}^\top\!\Vec{r}^{t+1}
\end{equation}
where $\mathbb{R}_+^{n \times m}\!\ni \!\Mat{K}\!=\!\textrm{exp}(\frac{-\Mat{D}^{p}}{\epsilon})$, \D{$t$ is current iteration number and $./$ denotes element-wise division for vectors. $\Vec{c}^0 \in \mathbb{R}_+^{m}$ is randomly initialized. A brief overview of the algorithm is presented in Algorithm~\ref{alg_sinkhorn}. Eqn.~\eqref{eqn:sinkhorn_iter} is repeated until $\Vec{r}^t$ and $\Vec{c}^t$ converges to their optimal values $\Vec{\tilde{r}}$ and $\Vec{\tilde{c}}$ respectively (as shown in Steps 8-12 of Algorithm~\ref{alg_sinkhorn}).}

\input{SOUMAVA-FILES/paper_algorithm}

\subsection*{Maximum Mean Discrepancy}
The Sinkhorn Divergence~\citep{genevay2018learning} between the probability distributions $\mathcal{U}$ and $\mathcal{V}$ is defined as follows:
\begin{equation}
    SD_\epsilon^p(\mathcal{U}, \mathcal{V}) = \mathcal{W}_\epsilon^{p}(\mathcal{U}, \mathcal{V}) - \frac{\mathcal{W}_\epsilon^{p}(\mathcal{U}, \mathcal{U}) + \mathcal{W}_\epsilon^{p}(\mathcal{V}, \mathcal{V}) }{2}
\end{equation}
It has been shown in~\citep{carlier2017convergence} that  $SD_{\epsilon\to0}^p(\mathcal{U}, \mathcal{V}) = \mathcal{W}^{p}(\mathcal{U}, \mathcal{V})$ (as defined in Eqn.~\eqref{eqn:discrete_ot}). Similarly the Energy Distance between $\mathcal{U}$ and $\mathcal{V}$ is defined as follows:
\begin{equation}
    \label{eqn:ed}
    \begin{aligned}
    ED(\mathcal{U}, \mathcal{V}) &= \mathcal{E}(\mathcal{U}, \mathcal{V}) - \frac{\mathcal{E}(\mathcal{U}, \mathcal{U}) + \mathcal{E}(\mathcal{V}, \mathcal{V}) }{2} \\
    &= \textrm{MMD}_{-\Mat{D}^{p}}(\mathcal{U}, \mathcal{V})
    \end{aligned}
\end{equation}
where $\mathcal{E}(\mathcal{U}, \mathcal{V})\!=\!\left \langle \Vec{w}\Vec{\widetilde{w}}^\top, \Mat{D}^p\right \rangle$, and $\left \langle \Vec{a}, \Vec{b} \right \rangle$ denotes the dot product between $\Vec{a}$ and $\Vec{b}$. 

Similarly in~\citep{genevay2018learning}, it is shown that $SD_{\epsilon\to\infty}^p(\mathcal{U}, \mathcal{V})=ED(\mathcal{U}, \mathcal{V})=\textrm{MMD}_{-\Mat{D}^{p}}(\mathcal{U}, \mathcal{V})$. $\textrm{MMD}_{-\Mat{D}^{p}}$ denotes the Maximum Mean Discrepancy~\citep{gretton2007kernel} between $\mathcal{U}$ and $\mathcal{V}$ with the kernel $k$ set to $-\Mat{D}^{p}$.~More generally, \textrm{MMD} provides an empirical estimate of the difference between $\mathcal{U}$ and $\mathcal{V}$ in a Reproducing Kernel Hilbert Space~\citep{alvarez2012kernels} $H$ as shown below:
\begin{equation} 
\label{eq:mmd}
    \begin{aligned}
    &\mathrm{MMD}(\mathcal{U},\mathcal{V})=\Big\|\frac{1} {n}\sum_{i=1}^n \Phi(\Mat{u}_i)-\frac{1} {m}\sum_{j=1}^m \Phi(\Mat{v}_j)\Big\|_H^2 \\
    &= \frac{1} {n^2}\sum_{i,i'}k(\Mat{u}_i,\Mat{u}_{i'}) \!-\!\frac{2}{nm}\sum_{i,j}k(\Mat{u}_i,\Mat{v}_j)\!+\!\frac{1} {m^2}\sum_{j,j'}k(\Mat{v}_j,\Mat{v}_{j'}) ~~,
    \end{aligned}
\end{equation}
where$~\Phi(\cdot)$ is a non-linear functional mapping that projects its input to a high-dimensional space.
\noindent In our experiments we have employed two different kernels, namely the (\RNum{1}) \underline{L}aplacian and (\RNum{2}) \underline{G}aussian kernels to calculate $\mathrm{MMD}$ between $\mathcal{U}$ and $\mathcal{V}$; which are shown below: 
\begin{equation}
\begin{split}
\label{eq:mmd_kernel}
    k_L(\Vec{u},\Vec{v})=\exp(-\frac{\|\Vec{u}-\Vec{v}\|}{\sigma}) ~ ; ~
    k_G(\Vec{u},\Vec{v})=\exp(-\frac{\|\Vec{u}-\Vec{v}\|^2_2}{2~\sigma^2}) ~.
\end{split}
\end{equation}

%% file: SOUMAVA-FILES/paper_algorithm.tex
\begin{algorithm}[!t]
\caption{Fixed point Sinkhorn Iterative Algorithm for discrete distributions.}
\label{alg_sinkhorn}
\begin{algorithmic}[1]
    \State Input: $\mathcal{U}$ and $\mathcal{V}$; and their respective support points $\{\Vec{u}_i\}_{i=1}^n$ and $\{\Vec{v}_j\}_{j=1}^m$.
    \State Input: The regularization parameter $\epsilon > 0$.
    \State Input: $\Vec{w}$ and $\Vec{\widetilde{w}}$ so as to satisfy Eqn.~\eqref{eqn:discretization}.
    \State Input: $\Mat{D}^{p}\in\mathbb{R}_+^{n \times m}$ where $\Mat{D}^{p}_{ij}=\textrm{d}(\Vec{u}_i, \Vec{v}_j)^{p}$ and $p \geq 1$.
    \algrule
    \State Output: $\mathrm{T}^{*}$ that satisfies Eqn.~\eqref{eqn:sinkhorn} s.t. $\mathrm{T}\Vec{1}=\Vec{w}$ and $\mathrm{T}^\top\Vec{1}=\Vec{\widetilde{w}}$. 
    \algrule \hspace{3cm}\textbf{Solution} \algrule
    \State $\Mat{K}\gets\textrm{exp}(\frac{-\Mat{D}^{p}}{\epsilon})$
    \State $t \gets 0$, $\Vec{c}^0 \gets \mathrm{R}_+^m$.
    \While {!\Big( $\left \| \Vec{r}^{t}-\Vec{r}^{t-1} \right \| \approx 0$ \&\& $\left \|  \Vec{c}^{t}-\Vec{c}^{t-1} \right \| \approx 0$\Big)} 
        \State $\Vec{r}^{t+1} \leftarrow \Vec{w} ./ \Mat{K} \Vec{c}^{t}$
        \State $\Vec{c}^{t+1} \leftarrow \Vec{\widetilde{w}} ./ \Mat{K}^\top\!\Vec{r}^{t+1}$
        \State $t \gets t+1$
    \EndWhile
    \State $\Vec{\tilde{r}} \gets \Vec{r}^{t}$, $\Vec{\tilde{c}} \gets \Vec{c}^{t}$, $\textrm{T}^*\!=\!\Delta(\Vec{\tilde{r}})~\textrm{exp}(\frac{-\Mat{D}^{p}}{\epsilon})~\Delta(\Vec{\tilde{c}})$
    \end{algorithmic}
\end{algorithm}

%% file: SOUMAVA-FILES/paper_ablation.tex
\begin{table}[!h]
    \centering
    \caption{Importance of $\mathrm{L}^{\mathrm{L}}$, $\mathrm{L}^{\mathrm{G}}$ and $\mathrm{L}^{\mathcal{W}}$ (Eqn.~(\textcolor{red}{5}) of the main text) 
    in clean and noisy labels settings.
    }
    \label{tab:ablation_1}
    \vspace{0.2cm}
    \scalebox{1}{
    \begin{tabular}{cc!{\vrule width 1.5pt}c!{\vrule width 1.5pt}c|c!{\vrule width 1.5pt}c|c!{\vrule width 1.5pt}}
            \cline{3-7}
            & & Clean & \multicolumn{2}{c!{\vrule width 1.5pt}}{Symmetric} & \multicolumn{2}{c!{\vrule width 1.5pt}}{Asymmetric} 
            \\
            \Xhline{2\arrayrulewidth}
            \multicolumn{1}{!{\vrule width 1.5pt}c|}{Dataset}& Method & $\delta$=$0.0$ & $\delta$=$0.1$ & $\delta$=$0.3$ & $\delta$=$0.1$ & $\delta$=$0.2$  \\
            \Xhline{2\arrayrulewidth}
            \multicolumn{1}{!{\vrule width 1.5pt}c|}{\multirow{3}{*}{S-$10$}}& $\mathrm{L}^{\mathrm{L}}$ & 42.8 & 40.5 & 33.6 & 41.0 & 40.3 \\
            \multicolumn{1}{!{\vrule width 1.5pt}c|}{} & $\mathrm{L}^{G}$ & 42.9 & 39.6 & 33.5 & 41.3 & 40.7 \\
            \multicolumn{1}{!{\vrule width 1.5pt}c|}{} & $\mathrm{L}^{\mathcal{W}}$ & 47.2 & 46.2 & 44.6 & 46.4 & 45.5  \\
            \Xhline{2\arrayrulewidth}
            \multicolumn{1}{!{\vrule width 1.5pt}c|}{\multirow{3}{*}{C-$10$}} & $\mathrm{L}^{\mathrm{L}}$ & 78.2 & 74.4 & 52.4 & 71.5 & 69.5 \\
            \multicolumn{1}{!{\vrule width 1.5pt}c|}{} & $\mathrm{L}^{\mathrm{G}}$ & 78.0 & 73.5 & 52.7 & 71.5 & 69.3 \\
            \multicolumn{1}{!{\vrule width 1.5pt}c|}{} & $\mathrm{L}^{\mathcal{W}}$ & 80.7 & 77.1 & 55.0 & 74.3 & 71.6 \\
            \Xhline{2\arrayrulewidth}
        \end{tabular}
    }
\end{table}

\section{Ablation Study}
\label{sec:eccv2020_ablation_study}

\par\noindent\textbf{Importance of $\mathrm{L}^{\mathrm{L}}$, $\mathrm{L}^{\mathrm{G}}$ and $\mathrm{L}^{\mathcal{W}}$:}
We first begin by providing a detailed insight into the effectiveness of using $\textrm{MMD}$ with Laplacian and Gaussian kernels, as well as the Wasserstein Distance in enforcing a class-wise discrepancy loss in both the clean and the noisy labels experimental setups. Specifically, we train our models with $\mathrm{L_{Train}}\!=\!\mathrm{L}^{\varphi}$; where $\mathrm{L}^{\varphi}\!=\!\left \{ \mathrm{L}^{\mathrm{L}}, \mathrm{L}^{\mathrm{G}}, \mathrm{L}^{\mathcal{W}} \right \}$ by discarding $\mathrm{L_{Local}}$ from Eqn.~(\textcolor{red}{5}) of the main text
and setting the value of $\lambda$ to $1.0$. The results for STL-$10$ and Cifar-$10$ datasets are reported in Table~\ref{tab:ablation_1}. It is observed that across all the settings; the Wasserstein Distance based discrepancy measure (\ie $\mathrm{L}^{\mathcal{W}}$) outperforms $\mathrm{MMD}$ based $\mathrm{L}^{\mathrm{L}}$ and $\mathrm{L}^{\mathrm{G}}$ in terms of classification accuracy, thus being able to exploit the underlying geometry to a better extent over $\mathrm{MMD}$, \C{unlike Wasserstein distances, the norm of MMD scales up as the batch size is increased with a low sample complexity~\citep{feydy2019interpolating}. Moreover, MMD generally seems to suffer from the drawback of vanishing gradient, while such difficulty rarely occurs in learning with Wasserstein distances~\citep{peyre2018computational}.}. Empirically, it is also observed in the majority of the cases that $\mathrm{L}^{\mathrm{L}}$ obtains similar or outperforms $\mathrm{L}^{\mathrm{G}}$ across both the datasets. Thus, from here onwards we only report the results of $\mathrm{MMD}$ with the Laplacian kernel (\ie $\mathrm{L^L}$).

\begin{table}[t]
    \centering
    \caption{Different values of $\lambda$ in the clean labels setting for S-$10$ dataset in the clean labels setting.
    }
    \label{tab:ablation_2}
    \scalebox{1.05}{
        \begin{tabular}{c|c|c|c|c|c|c|c|c|c!{\vrule width 1.5pt}}
            \cline{2-10}
            & $\lambda$ & 0.0 & 0.01 & 0.1 & 0.2 & 0.5 & 0.7 & 1.0 & 5.0 
            \\ \Xhline{2\arrayrulewidth}
            \multicolumn{1}{!{\vrule width 1.5pt}c|}{\multirow{2}{*}{$\mathrm{L_{Trip}}$}} & $+\mathrm{L}^{\mathrm{L}}$ & \multirow{2}{*}{59.9} & 59.4 & 60.9 & \textbf{61.6} & 61.2 & 61.6 & 60.8 & 58.5 
            \\ \cline{2-2} \cline{4-10} 
            \multicolumn{1}{!{\vrule width 1.5pt}c|}{} & $+\mathrm{L}^{\mathcal{W}}$ &  & 64.1 & 63.8 & 66.1 & \textbf{66.1} & 65.8 & 65.3 & 63.6 \\ \Xhline{2\arrayrulewidth}
            \multicolumn{1}{!{\vrule width 1.5pt}c|}{\multirow{2}{*}{$\mathrm{L_{NP}}$}} & $+\mathrm{L}^{\mathrm{L}}$ &  \multirow{2}{*}{44.8} & 42.7 & 45.7 & \textbf{45.4} & 43.4 & 43.1 & 42.9 & 41.9 
            \\ \cline{2-2} \cline{4-10} 
            \multicolumn{1}{!{\vrule width 1.5pt}c|}{} & $+\mathrm{L}^{\mathcal{W}}$ & & 46.9 & 48.5 & 49.7 & \textbf{49.8} & 49.4 & 48.3 & 45.5
            \\ \Xhline{2\arrayrulewidth}
        \end{tabular}
        }
\end{table}

\par\noindent\textbf{Importance of $\lambda$:} Table~\ref{tab:ablation_2} shows the classification results obtained for various values of $\lambda$ on STL-$10$ dataset in the clean labels setting. The optimal value of $\lambda$ is $0.2$ and $0.5$ for  $\mathrm{L^{L}}$ and $\mathrm{L^{\mathcal{W}}}$ respectively; which have been chosen to report the results in the subsequent sections. One can observe that the classification accuracy increases when $\lambda$ is increased from $0.01$ with the optimum accuracy being attained at their respective chosen values of $\lambda$. Thereafter it drops when $\lambda$ is increased beyond $1.0$. One can also see improvements over $\mathrm{L_{Trip}}$ and $\mathrm{L_{NP}}$ (especially for $\mathrm{L^{\mathcal{W}}}$) even when the value $\lambda$ is outside its optimal range. \D{There also exists a large performance between $\mathrm{L_{Trip}}$ and $\mathrm{L_{NP}}$ for $\lambda$=0. One plausible explanation is that $\mathrm{L_{NP}}$ considers all the negative samples for a given pair of anchor and positive sample in calculating $\mathrm{L_{NP}}$ (Refer to Section~\textsection~\textcolor{red}{3} of the main text 
for more details regarding $\mathrm{L_{NP}}$). As a result, some of these negative samples might be detrimental to the optimization procedure as they might be too difficult to separate given the anchor-positive pair. Further, $\mathrm{L_{NP}}$ can form only one fixed positive pair per anchor sample within the mini-batch, thereby resulting in less variability in the positive pair of samples so as to overcome the adverse effects caused by the presence of large number of negative samples within the training batch. On the other hand, $\mathrm{L_{Trip}}$ enforces semi-hard triplet mining to select the most informative triplets in training the model; thereby outperforming $\mathrm{L_{NP}}$ by a significant margin.}

\par\noindent\textbf{Embedding Size:} We have performed experiments with different dimensions of the embedding space on the STL-10 dataset with $\mathrm{L_{Local}}$ set to $\mathrm{L_{Trip}}$ in the clean labels setting. When trained with $\mathrm{L}^{\mathrm{L}}$, we obtain $61.1\%$, $61.6\%$, $61.7\%$ and $60.5\%$ accuracy for $32$, $64$, $128$ and $256$ dimensional embedding space respectively. Similarly for $\mathrm{L}^{\mathcal{W}}$, we observed $64.2\%$, $66.1\%$, $64.7\%$ and $64.3\%$ accuracy respectively. This verifies that a $64$ dimensional embedding space obtains superior results for $\mathrm{L}^{\mathcal{W}}$, while achieving competitive performance against a $128$ dimensional embedding space for $\mathrm{L}^{\mathrm{L}}$. We have therefore set the dimensionality of the embedding space to $64$ for all the datasets in the forthcoming sections.

\par\noindent\textbf{More samples per class:} We have also conducted experiments with $1000$ samples for STL-$10$ dataset (\ie $100$ samples per class) and obtained $63.7\%$ accuracy on its test set against $66.1\%$ when trained with 100 samples (\ie $10$ samples per class) per mini-batch respectively with $\mathrm{L_{Local}}$ set to $\mathrm{L_{Trip}}$ in the clean labels setting. Similar to~\citep{shen2018wasserstein}, this study clearly demonstrates that $\textrm{DCDL}$ is able to approximate the underlying distribution with fewer samples per mini-batch. In general, assuming that larger batches should always lead to better results does not hold as shown in~\citep{golmant2018computational,masters2018revisiting}. One intuitive way of understanding this is to think of gradients of small batches as noisy gradients which can help escaping from bad local minima.

\input{SOUMAVA-FILES/paper_complexity}

%% file: SOUMAVA-FILES/paper_complexity.tex
\par\noindent\textbf{Computational complexity:} The computational complexity of the regularized Sinkhorn Divergence (SD) is $\mathcal{O}(\textrm{N}_{\mathrm{B}}^2)$, where $\textrm{N}_{\mathrm{B}}$ is the number of samples in a mini-batch. However, a big portion of the computational training time is devoted to computing the pair-wise distances between samples. Such pairwise distances are calculated in $\mathrm{L_{Local}}$ which can be used explicitly in the SD module; thereby reducing its overall computational overhead. As observed, the average computational time per training epoch is 0.53, 0.57 and 0.61 seconds for $\mathrm{L_{Trip}}$ and its MMD (\ie $\mathrm{L_{Trip}}+\mathrm{L}^{\mathrm{L}}$) and Wasserstein Distance (\ie $\mathrm{L_{Trip}}+\mathrm{L}^{\mathcal{W}}$) counterparts respectively when trained on the STL-10 dataset with a $64$ dimensional embedding space in the clean labels settings. This clearly shows that the additional computational overhead of SD is very small due to the sharing of pairwise distance calculations among $\mathrm{L_{Local}}$ and $\mathrm{L}^{\varphi}$, while the inference time remains the same.

%% file: SOUMAVA-FILES/paper_clean_labels.tex
\begin{table}[t]
    \centering
    \caption{Experimental results demonstrating the importance of incorporating $\mathrm{L}^{\mathrm{L}}$ and $\mathrm{L}^{\mathcal{W}}$ in Eqn.~(\textcolor{red}{5}) (of the main text) across the three datasets in the \textbf{clean} labels setting.
    }
    \label{tab:clean}
    \vspace{0.2cm}
    \scalebox{1.1}{
    \begin{tabular}{!{\vrule width 1.5pt}c!{\vrule width 1.5pt}c|c|c!{\vrule width 1.5pt}}
    \Xhline{2\arrayrulewidth}
       Dataset & C-10 & C-100 & S-10  \\
       \Xhline{2\arrayrulewidth}
        $\mathrm{L_{Trip}}$ & 85.9 & 68.4 & 59.9 \\
        $\mathrm{L_{Trip}}$ + $\mathrm{L}^{\mathrm{L}}$ & 87.5 & 69.2 & 61.6 \\
        $\mathrm{L_{Trip}}$ + $\mathrm{L}^{\mathcal{W}}$ & 87.4 & 70.1 & 66.1 \\
        \Xhline{2\arrayrulewidth}
        $\mathrm{L_{NP}}$ & 68.6 & 42.6 & 44.8 \\
        $\mathrm{L_{NP}}$ + $\mathrm{L}^{\mathrm{L}}$ & 69.9 & 44.6 & 45.4 \\
        $\mathrm{L_{NP}}$ + $\mathrm{L}^{\mathcal{W}}$ & 70.2 & 45.5 &  49.8 \\
        \Xhline{2\arrayrulewidth}
        $\mathrm{L_{Ang}}$  & 85.7 & 63.0 & 54.6  \\
        $\mathrm{L_{Ang}}$ + $\mathrm{L}^{\mathrm{L}}$ & 86.6 & 63.6 & 57.0 \\
        $\mathrm{L_{Ang}}$ + $\mathrm{L}^{\mathcal{W}}$ & 87.2 & 65.2 &  59.4 \\
        \Xhline{2\arrayrulewidth}
        $\mathrm{L_{Ang\_NP}}$  & 84.5 & 61.6 & 53.4 \\
        $\mathrm{L_{Ang\_NP}}$ + $\mathrm{L}^{\mathrm{L}}$ & 85.3 & 63.6 & 55.9 \\
        $\mathrm{L_{Ang\_NP}}$ + $\mathrm{L}^{\mathcal{W}}$ & 87.1 & 66.1 & 57.4 
    \\\Xhline{2\arrayrulewidth}
    \end{tabular}
    }
\end{table}

\section{Clean Labels:}
Here, we assess the effectiveness of our proposed algorithm~$\mathrm{DCDL}$ in the clean labels settings. The results on the three datasets are shown in Table~\ref{tab:clean} ($\delta$ set to $0.0$). As observed, enforcing discrepancy between the distributions of the class-wise embeddings leads to an enhancement in the discriminative ability of the learnt embedding space. Further, $\mathrm{L}^{\mathcal{W}}$ achieves significant performance gains over $\mathrm{L}^{\mathrm{L}}$
for different assignments of $\mathrm{L_{Local}}$ losses (Refer to Eqn.~(\textcolor{red}{5}) of the main text).
Thus it is indeed evident that one must enforce a 
class-wise discrepancy loss that aims to separate the probability distributions of the embedded points for every class in order to learn well-separated, yet compact clusters in the embedding space.

%% file: SOUMAVA-FILES/paper_fgor.tex
\section{Fine Grained Image Classification}
\label{sec:eccv2020_fgor}
\begin{figure*}
	\centering
	\label{fig:fgor}
	\subfigure[CUBS200-2011~\cite{CUB200_DB}]{\includegraphics[width=0.45\textwidth,keepaspectratio]{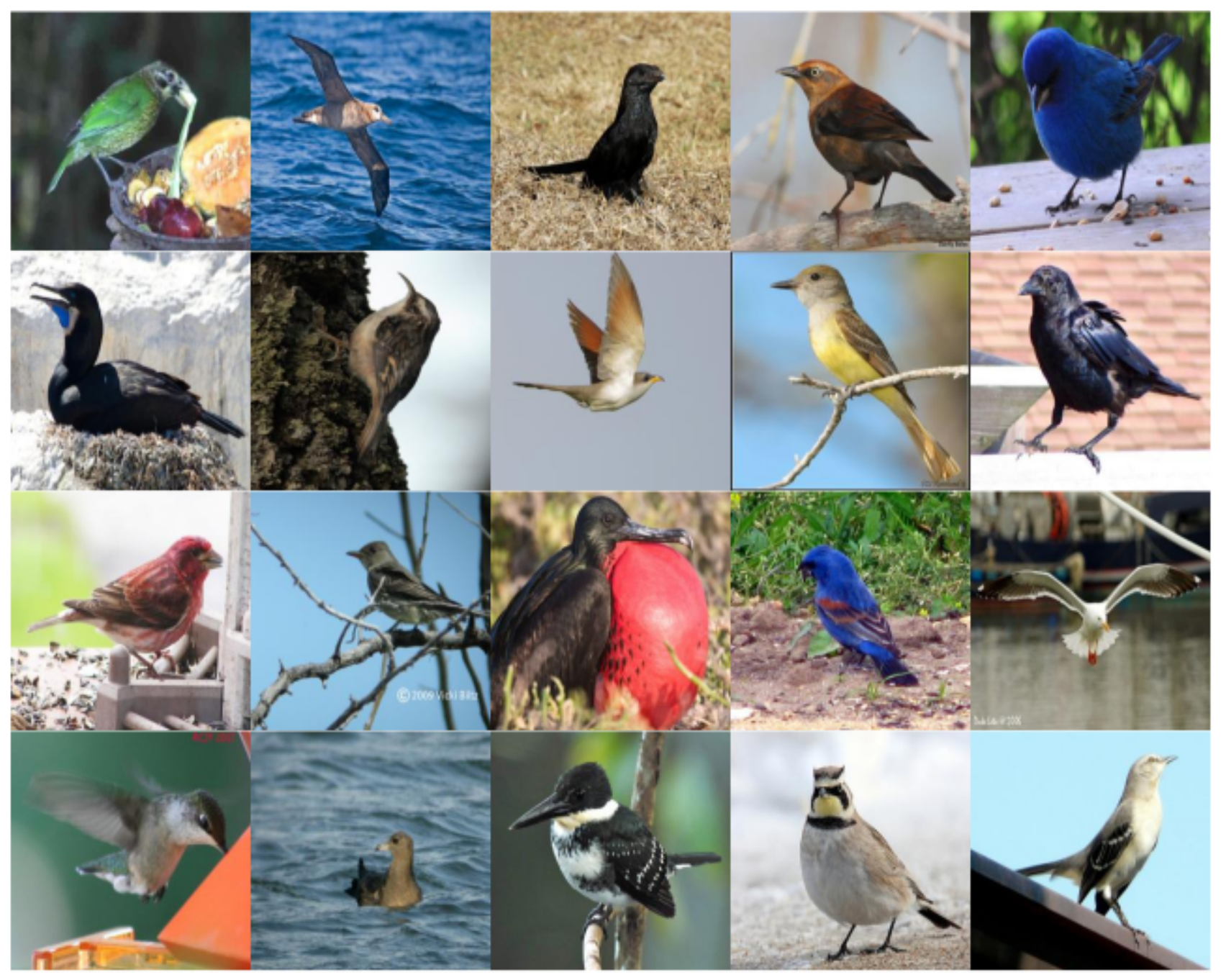}\label{fig:cubs}}%
	\hfil
	\subfigure[CARS196~\cite{CARS196_DB}]{\includegraphics[width=0.45\textwidth,keepaspectratio]{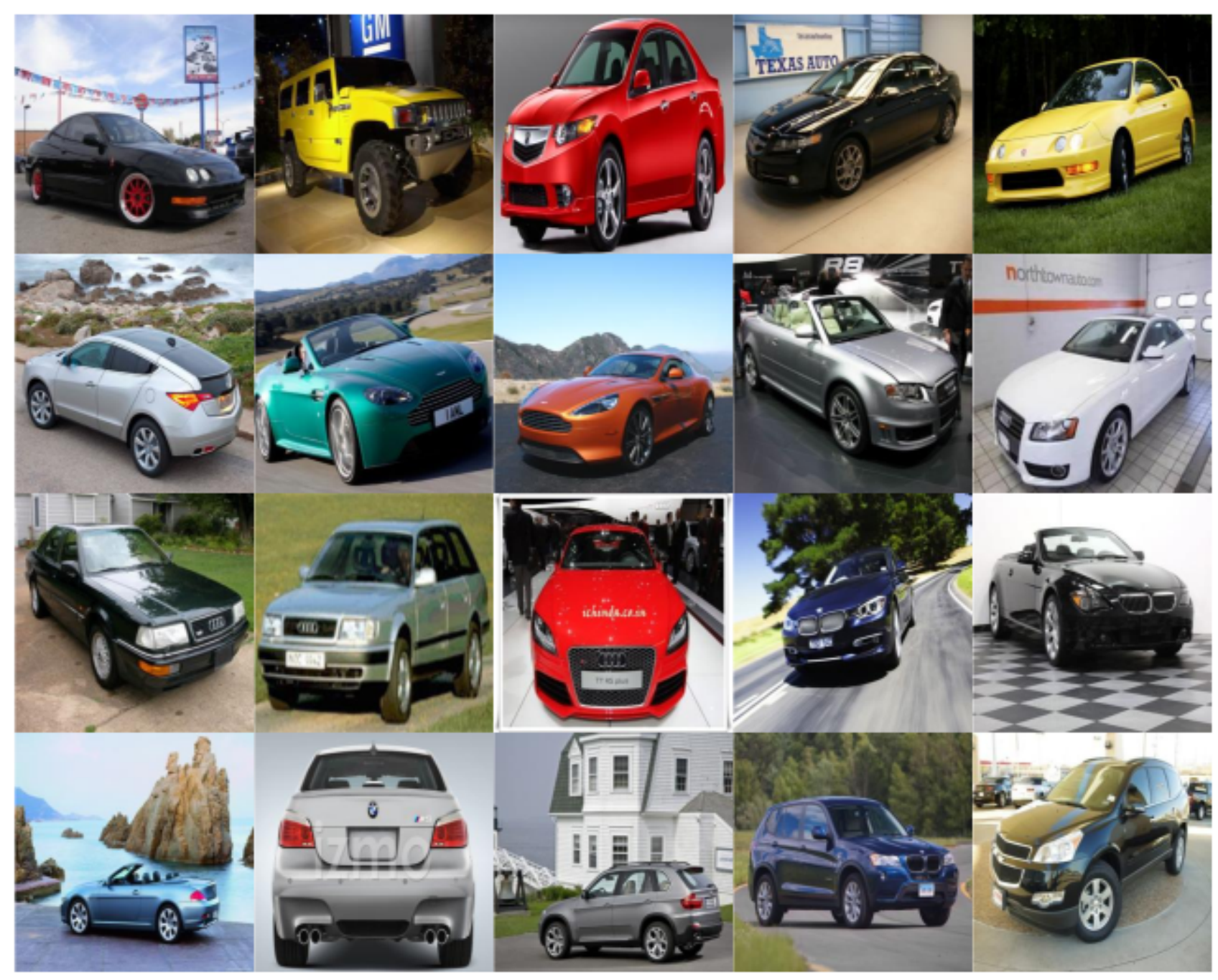}\label{fig:cars}}
	\vspace{0.2cm}
	\caption{Exemplar images from the fine-grained image recognition datasets.}
 \end{figure*}

We also evaluate the $\mathrm{DCDL}$ method on the widely used fine-grained image datasets  \textbf{(a) Caltech-UCSD Birds} (CUBS-$200$-$2011$)~\citep{CUB200_DB} and \textbf{(b) Stanford Cars} dataset (CARS$196$)~\citep{CARS196_DB}. A brief description of the datasets is provided below:

\begin{itemize}
    \item The \emph{Caltech-UCSD Birds} (\textbf{CUBS-$200$-$2011$)}~\citep{CUB200_DB} consists of 11,788 images of birds from 200 different varieties. The first 100 categories are considered for training (5,864 images), while the rest 100 categories are considered for testing (5,924 images).
    
    \item The \emph{Stanford Cars} (\textbf{CARS196}) dataset~\citep{CARS196_DB} consists of 16,185 images of cars from 196 different categories. The first 98 categories (8,054 images) are considered in the training of the network, while the next 98 categories (8,131 images) are used in the testing phase. 
\end{itemize} Exemplar images of CUBS-200-2011~\cite{CUB200_DB} and CARS196~\cite{CARS196_DB} are shown in Fig.~\ref{fig:cubs} and~\ref{fig:cars} respectively.

\section*{Implementation Details:} We have used Inception-V1 \citep{szegedy2015going} with Batch Normalization \citep{ioffe2015batch} pretrained on Imagenet~\citep{russakovsky2015imagenet} as the backbone network, with a randomly initialized~\citep{glorot2010understanding} fully-connected embedding layer at the end\footnote{Like~\citep{Law_ICML_2017}, we also pre-train the entire network with a classification loss before fine-tuning}. Similar to~\citep{Law_ICML_2017}, we have fixed the dimensionality of the embedding to $100$ and $98$ respectively for CUBS-$200$-$2011$ and CARS$196$. All the images are resized to $256 \times 256$ initially. The training images are then randomly cropped to $227 \times 227$, followed by random horizontal flipping; whereas the test images are cropped to $227 \times 227$ from the center. A Dropout layer with dropout rate of $0.1$ is added before the embedding layer. The size of the mini-batch is set to $64$ for both datasets. In our experiments with $\mathrm{L_{Local}}\!=\!\mathrm{L_{Trip}}$, we ensure that there are at-least $4$ samples per class within the mini-batch $\mathrm{B}$; while for the rest we use $2$ samples per class in order to create the mini-batch. The entire network is fined tuned using the Stochastic Gradient Descent (SGD) optimizer. The initial learning rate is set to $0.01$ for all datasets, and is decreased by a factor of $0.1$ after every $25$ epochs. 
However, empirically we have found that $\lambda\!\!=\!\!0.5$ gives superior results for experiments with $\mathrm{L}^{\mathrm{L}}$ and $\mathrm{L}^{\mathcal{W}}$. Moreover, the value of $\epsilon$ is also set to $2.5\times10^{-3}$. We set the value of \textbf{(a)} $\tau$ in $\mathrm{L_{Trip}}$ to $0.5$, \textbf{(b)} $\alpha$ in $\mathrm{L_{Ang}}$ and $\mathrm{L_{Ang\_NP}}$ to $\ang{45}$. We report the best results obtained for all the hyper-parameter settings after training the models for $40$ epochs.

\begin{table*}[!t]
\caption{Experimental results on CUBS-$200$-$2011$ in the \textbf{clean} and \textbf{symmetric}-noisy label settings.}
\centering
\label{tab:fgor_cubs}
\vspace{0.2cm}
\scalebox{0.85}{
\begin{tabular}{!{\vrule width 1.5pt}c|c|c|c|c||c|c|c|c|c!{\vrule width 1.5pt}}
    \Xhline{3\arrayrulewidth} 
    & \multicolumn{9}{c!{\vrule width 1.5pt}}{Clean Labels} \\ 
    \Xhline{3\arrayrulewidth}
    Method & NMI & R@1 & R@2 & R@4 & Method & NMI & R@1 & R@2 & R@4 \\ \Xhline{2\arrayrulewidth}
    $\mathrm{L_{Trip}}$& 55.4 & 42.6 & 55.1 & 66.4 & $\mathrm{L_{NP}}$ & 57.2 & 45.4 & 58.4 & 69.5 \\ 
$\mathrm{L_{Trip}}$ + $\mathrm{L}^{\mathrm{L}}$ & 56.2 & 43.3 & 55.2 & 66.8 & $\mathrm{L_{NP}}$ + $\mathrm{L}^{\mathrm{L}}$ & 57.4 & 45.3 & 58.3 & 70.6 \\ 
$\mathrm{L_{Trip}}$ + $\mathrm{L}^{\mathcal{W}}$ & 56.9 & 44.1 & 56.1 & 67.5 & $\mathrm{L_{NP}}$ + $\mathrm{L}^{\mathcal{W}}$ & 57.9 & 45.9 & 58.5 & 70.5\\
 \hline
    $\mathrm{L_{Ang}}$                               & 57.7 & 45.7 & 58.1 & 70.3 & $\mathrm{L_{Ang\_NP}}$ & 57.9 & 45.9 & 58.3 & 70.4 \\ 
    $\mathrm{L_{Ang}}$ + $\mathrm{L}^{\mathrm{L}}$            & 58.1 & 46.4 & 58.7 & 70.8  & $\mathrm{L_{Ang\_NP}}$ + $\mathrm{L}^{\mathrm{L}}$ & 58.1 & 46.4 & 58.7 & 70.5 \\
    $\mathrm{L_{Ang}}$ + $\mathrm{L}^{\mathcal{W}}$  & 58.5 & 46.3 & 58.7 & 70.7 & $\mathrm{L_{Ang\_NP}}$ + $\mathrm{L}^{\mathcal{W}}$ & 58.3 & 46.2 & 58.5 & 70.4 \\
     \Xhline{3\arrayrulewidth}
     & \multicolumn{9}{c!{\vrule width 1.5pt}}{Symmetric Noise with $\delta$=0.1} \\ \cline{1-10} 
$\mathrm{L_{Trip}}$& 53.9 & 41.7 & 53.8 & 64.5 &$\mathrm{L_{NP}}$ & 55.2 & 43.7 & 56.3 & 68.3 \\
$\mathrm{L_{Trip}}$ + $\mathrm{L}^{\mathrm{L}}$ & 54.3 & 42.3 & 54.1 & 65.0 & $\mathrm{L_{NP}}$ + $\mathrm{L}^{\mathrm{L}}$ & 56.9 & 45.2 & 57.5 & 68.5 \\
$\mathrm{L_{Trip}}$ + $\mathrm{L}^{\mathcal{W}}$ & 54.7 & 42.5 & 54.4 & 65.4 & $\mathrm{L_{NP}}$ + $\mathrm{L}^{\mathcal{W}}$ & 57.3 & 44.8 & 57.9 & 69.8 \\
\hline
    $\mathrm{L_{Ang}}$                               & 56.9 & 45.0 & 57.1 & 69.5 & $\mathrm{L_{Ang\_NP}}$ + $\mathrm{L}^{\mathrm{L}}$ & 57.2 & 45.7 & 58.7 & 70.1 \\
    $\mathrm{L_{Ang}}$ + $\mathrm{L}^{\mathrm{L}}$            & 57.4 & 45.4 & 58.0 & 69.8 & $\mathrm{L_{Ang\_NP}}$ + $\mathrm{L}^{\mathrm{L}}$ & 57.9 & 45.6 & 57.8 & 69.8 \\
    $\mathrm{L_{Ang}}$ + $\mathrm{L}^{\mathcal{W}}$  & 58.0 & 45.5 & 58.1 & 70.4 & $\mathrm{L_{Ang\_NP}}$ + $\mathrm{L}^{\mathcal{W}}$ & 57.8 & 45.7 & 56.9 & 69.6 \\
    \Xhline{3\arrayrulewidth}
    & \multicolumn{9}{c!{\vrule width 1.5pt}}{Symmetric Noise with $\delta$=0.3} \\ \cline{1-10}
    $\mathrm{L_{Trip}}$& 52.1 & 40.8 & 52.6 & 62.8 & $\mathrm{L_{NP}}$ & 54.7 & 42.0 & 54.6 & 66.6 \\
$\mathrm{L_{Trip}}$ + $\mathrm{L}^{\mathrm{L}}$ & 52.9 & 41.4 & 53.1 & 62.6 & $\mathrm{L_{NP}}$ + $\mathrm{L}^{\mathrm{L}}$ & 55.9 & 43.1 & 55.3 & 67.2 \\
$\mathrm{L_{Trip}}$ + $\mathrm{L}^{\mathcal{W}}$ & 53.7 & 42.2 & 54.0 & 63.9 & $\mathrm{L_{NP}}$ + $\mathrm{L}^{\mathcal{W}}$ & 56.1 & 43.8 & 56.3 & 67.8 \\
\hline
    $\mathrm{L_{Ang}}$                             & 55.4 & 42.1 & 55.1 & 67.3 & $\mathrm{L_{Ang\_NP}}$ & 56.5 & 42.9 & 55.9 & 68.0 \\ 
    $\mathrm{L_{Ang}}$ + $\mathrm{L}^{\mathrm{L}}$            & 56.2 & 42.8 & 56.0 & 68.0 & $\mathrm{L_{Ang\_NP}}$ + $\mathrm{L}^{\mathrm{L}}$ & 56.1 & 43.5 & 56.1 & 67.9 \\
    $\mathrm{L_{Ang}}$ + $\mathrm{L}^{\mathcal{W}}$  & 56.5 & 43.0 & 55.9 & 68.2 & $\mathrm{L_{Ang\_NP}}$ + $\mathrm{L}^{\mathcal{W}}$ & 56.2 & 43.7 & 56.3 & 68.6 \\
    \Xhline{3\arrayrulewidth} 
\end{tabular}}
\end{table*}

In the evaluation phase, we apply the standard \textrm{KMeans} algorithm on the unit-norm output representations and calculate the conventional NMI and R@K metrics similar to~\citep{Song_CVPR_2016,Sohn_NIPS_2016,Song_CVPR_2017}. Like before, we evaluate $\mathrm{DCDL}$ in two different label settings: \textbf{(a) clean} labels and \textbf{(b) noisy} labels; with symmetric noise ($\delta$=$\{0.1, 0.3\}$). 

\begin{table*}[!t]
\centering
\caption{Experimental results on CARS196 dataset in the \textbf{clean} and \textbf{symmetric}-noisy label settings.}
\centering
\label{tab:fgor_cars}
\vspace{0.2cm}
\scalebox{0.85}{
\begin{tabular}{!{\vrule width 1.5pt}c|c|c|c|c||c|c|c|c|c!{\vrule width 1.5pt}}
    \Xhline{3\arrayrulewidth} 
     & \multicolumn{9}{c!{\vrule width 1.5pt}}{Clean Labels} \\ 
     \Xhline{3\arrayrulewidth}
     & NMI & R@1 & R@2 & R@4 & & NMI & R@1 & R@2 & R@4 \\ \Xhline{2\arrayrulewidth}
    $\mathrm{L_{Trip}}$                 & 55.4 & 58.1 & 69.8 & 79.5 &  $\mathrm{L_{NP}}$ & 56.9 & 60.8 & 72.5 & 80.9\\ 
    $\mathrm{L_{Trip}}$ + $\mathrm{L}^{\mathrm{L}}$ & 55.9 & 58.9 & 70.7 & 80.5 & $\mathrm{L_{NP}}$ + $\mathrm{L}^{\mathrm{L}}$& 57.3 & 62.1 & 73.2 & 81.7 \\ 
    $\mathrm{L_{Trip}}$ + $\mathrm{L}^{\mathcal{W}}$ & 56.3 & 58.6 & 70.8 & 80.9 & $\mathrm{L_{NP}}$ + $\mathrm{L}^{\mathcal{W}}$ & 57.2 & 62.0 & 73.3 & 81.6 \\ 
    \hline
    $\mathrm{L_{Ang}}$                             & 57.1 & 60.9 & 71.7 & 81.0 & $\mathrm{L_{Ang\_NP}}$ & 57.1 & 61.2 & 72.4 & 81.2 \\ 
    $\mathrm{L_{Ang}}$ + $\mathrm{L}^{\mathrm{L}}$            & 57.5 & 61.5 & 72.9 & 81.5 & $\mathrm{L_{Ang\_NP}}$ + $\mathrm{L}^{\mathrm{L}}$ &  57.4 & 61.9 & 72.9 & 81.7 \\
    $\mathrm{L_{Ang}}$ + $\mathrm{L}^{\mathcal{W}}$  & 57.9 & 61.8 & 72.8 & 81.2 & $\mathrm{L_{Ang\_NP}}$ + $\mathrm{L}^{\mathcal{W}}$ & 57.3 & 61.8 & 72.8 & 81.6 \\
     \Xhline{3\arrayrulewidth}
     & \multicolumn{9}{c!{\vrule width 1.5pt}}{Symmetric Noise with $\delta$=0.1} \\ \cline{1-10} 
     
    $\mathrm{L_{Trip}}$                 & 54.5 & 58.5 & 69.6 & 79.4 &  $\mathrm{L_{NP}}$ & 54.7 & 58.6 & 70.5 & 79.9 \\ 
    $\mathrm{L_{Trip}}$ + $\mathrm{L}^{\mathrm{L}}$  & 54.8 & 58.7 & 70.4 & 79.7 & $\mathrm{L_{NP}}$ + $\mathrm{L}^{\mathrm{L}}$& 55.5 & 59.1 & 70.6 & 79.7 \\ 
    $\mathrm{L_{Trip}}$ + $\mathrm{L}^{\mathcal{W}}$ & 55.3 & 58.9 & 70.4 & 80.2 & $\mathrm{L_{NP}}$ + $\mathrm{L}^{\mathcal{W}}$ & 55.4 & 59.2 & 71.1 & 80.1 \\ 
    \hline
    $\mathrm{L_{Ang}}$                             & 54.2 & 58.3 & 70.3 & 79.2 & $\mathrm{L_{Ang\_NP}}$ & 54.7 & 58.2 & 70.1 & 79.4 \\ 
    $\mathrm{L_{Ang}}$ + $\mathrm{L}^{\mathrm{L}}$            & 55.0 & 58.3 & 70.0 & 79.3 & $\mathrm{L_{Ang\_NP}}$ + $\mathrm{L}^{\mathrm{L}}$ & 55.2 & 59.3 & 71.4 & 80.5 \\
    $\mathrm{L_{Ang}}$ + $\mathrm{L}^{\mathcal{W}}$  & 54.7 & 59.5 & 70.8 & 80.0 & $\mathrm{L_{Ang\_NP}}$ + $\mathrm{L}^{\mathcal{W}}$ & 55.6 & 59.6 & 70.7 & 80.2 \\
    \Xhline{3\arrayrulewidth}
    
    & \multicolumn{9}{c!{\vrule width 1.5pt}}{Symmetric Noise with $\delta$=0.3} \\ \cline{1-10}
    $\mathrm{L_{Trip}}$                 & 49.3  & 51.1 & 62.8 & 73.8 & $\mathrm{L_{NP}}$ & 49.2 & 50.8 & 62.6 & 73.6 \\ 
    $\mathrm{L_{Trip}}$ + $\mathrm{L}^{\mathrm{L}}$ & 49.6 & 52.1 & 64.1 & 75.4 & $\mathrm{L_{NP}}$ + $\mathrm{L}^{\mathrm{L}}$& 49.7 & 51.1 & 63.4 & 74.3 \\ 
    $\mathrm{L_{Trip}}$ + $\mathrm{L}^{\mathcal{W}}$ & 49.3 & 52.1 & 64.0 & 74.9 & $\mathrm{L_{NP}}$ + $\mathrm{L}^{\mathcal{W}}$ & 49.9 & 51.9 & 64.3 & 74.9 \\ 
    \hline
    $\mathrm{L_{Ang}}$                             &  50.1 & 51.4 & 63.2 & 73.9 & $\mathrm{L_{Ang\_NP}}$ & 49.5 & 50.8 & 63.2 & 73.9 \\ 
    $\mathrm{L_{Ang}}$ + $\mathrm{L}^{\mathrm{L}}$            & 50.2 & 51.3 & 63.7 & 74.7 & $\mathrm{L_{Ang\_NP}}$ + $\mathrm{L}^{\mathrm{L}}$ & 50.1 & 51.6 & 63.7 & 74.6 \\
    $\mathrm{L_{Ang}}$ + $\mathrm{L}^{\mathcal{W}}$  & 50.3 & 52.2 & 64.5 & 74.9 & $\mathrm{L_{Ang\_NP}}$ + $\mathrm{L}^{\mathcal{W}}$ & 49.9 & 51.4 & 64.2 & 74.4 \\
    \Xhline{3\arrayrulewidth} 
\end{tabular}}
\end{table*}

Table~\ref{tab:fgor_cubs} and \ref{tab:fgor_cars} shows the results with $\mathrm{L_{Local}}\!= \!\left \{ \mathrm{L_{Trip}},~\mathrm{L_{NP}},~\mathrm{L_{Ang}},~ \mathrm{L_{Ang\_NP}} \right \}$ for the clean and symmetric-noisy labels settings on the CUBS-$200$-$2011$ and CARS$196$ datasets respectively. It is again evident that adding both $\mathrm{L}^{\mathrm{L}}$ and $\mathrm{L}^{\mathcal{W}}$ (so as to introduce class-wise discrepancy constraints) leads to superior NMI and R@K metrics for both the datasets across most of the settings (\ie clean and noisy labels). It further reinforces our hypothesis of the need to incorporate and enforce such class-wise discrepancy constraints in order to learn a discriminative embedding space. Furthermore, DSC~\citep{Law_ICML_2017} designs a complex structured loss that takes into consideration the spectral clustering constraints to separate the class-wise embeddings and achieves $56.1\%/57.1\%$ (NMI/R@1) on the CARS196 dataset. On the other hand, simply by precisely enforcing the discrepancy constraints with either a simpler $\mathrm{L_{NP}}$, $\mathrm{L_{Ang}}$, or $\mathrm{L_{Ang\_NP}}$  outperforms DSC in terms of NMI/R@1 respectively for both $\mathrm{L}^{\mathrm{L}}$ and $\mathrm{L}^{\mathcal{W}}$. These results clearly indicate that modeling comprehensive class-wise discrepancy constraints using $\mathrm{L}^{\mathrm{L}}$ and $\mathrm{L}^{\mathcal{W}}$ to \textbf{push-away} the class-wise probability distributions are indeed important to learn a discriminative embedding space in the presence/absence of noisy labels. 

\begin{table*}[t]
\centering
\caption{Comparison against several baselines and state-of-the-art algorithms. The best results are shown in \textcolor{red}{red}.}
\label{tab:sota}
\vspace{0.2cm}
\scalebox{1.0}{
\begin{tabular}{!{\vrule width 1.5pt}c!{\vrule width 1.5pt}c|c|c|c!{\vrule width 1.5pt}c|c|c|c!{\vrule width 1.5pt}}
\Xhline{3\arrayrulewidth}
Dataset & \multicolumn{4}{c!{\vrule width 1.5pt}}{CUBS-$200$-$2011$} & \multicolumn{4}{c!{\vrule width 1.5pt}}{CARS$196$} \\ \Xhline{3\arrayrulewidth}
Method & NMI & R@1 & R@2 & R@4 & NMI & R@1 & R@2 & R@4 \\ \Xhline{2\arrayrulewidth}
Trip & 55.4 & 42.6 & 54.9 & 66.4 & 53.4  & 51.5  & 63.8  & 73.5 \\ \hline
DSC & 58.1  & 49.8  & 62.6  & 73.6  & 58.0  & 59.4  & 71.3  & 80.6 \\ \hline
Lifetd-Struct & 56.5 & 43.6 & 56.6 & 68.6 & 56.9 & 53.0 & 65.7 & 76.0  \\ \hline
NPairs & 57.2 & 45.4 & 58.4 & 69.5 & 57.8 & 53.9 & 66.8 & 77.8  \\ \hline
NMI-based & 59.2 & 48.2 & 61.4 & 71.8 & 59.0 & 58.1 & 70.6 & 80.3 \\ \hline
Stiefel & 62.3 & 52.3 & 64.5 & 75.3 & 64.2 & 73.2 & 82.2 & 88.6 \\ \hline
Hist & - & 52.8 & 64.4 & 74.7 & - & 66.2 & 77.2 & 85.0\\ \hline 
Ang & 61.0 & 53.6 & 65.0 & 75.3 & 62.7 & 68.9 & 78.9 & 85.8 \\ \hline
DAML & 61.3 & 52.7 & 65.4 & 75.5 & 66.0 & 75.1 & 83.8 & 89.7 \\ \hline
HADML & 62.6 & 53.7 & 65.7 & 76.7 &  69.7 & 79.1 & 87.1 & 92.1 \\ \hline
DVML & 61.4 & 52.7 & 65.1 & 75.5 & 67.6 & 82.0 & 88.4 & \textcolor{red}{93.3} \\ \hline
BIER & - & 55.3 & 67.2 & 76.9 & - & 78.0 & 85.8 & 91.1 \\ \hline
HTL & - & 57.1 & 68.8 & 78.7 & - & 81.4 & 88.0 & 92.7 \\ \hline
HTG & - & 59.5 & 71.8 & 81.3 & - & 76.5 & 84.7 & 90.4 \\ \hline
A-BIER & - & 57.5 & 68.7 & 78.3 & - & 82.0 & 89.0 & 93.2 \\ \hline
DWS & 66.3 & 62.9 & 74.1 & 82.9 & 66.3 & 80.0 & 87.7 & 92.3\\ \hline
ProxyNCA & 62.5 & 57.4 & 69.2 & 79.1 & 59.5 & 73.0 & 81.3 & 87.9 \\ \hline
MIC & 69.7 &  66.1 & 76.8 & \textcolor{red}{85.6} & 68.4 & \textcolor{red}{82.6} & \textcolor{red}{89.1} & 93.2 \\ \hline
MIC$^\ast$ &  69.1 & 64.9 & 75.8 & 84.7 & 68.1 & 80.7 & 88.0 & 92.9 \\ \Xhline{3\arrayrulewidth} 
MIC+$\mathrm{L}^{\mathrm{L}}$ & 70.1 & 65.8 & 77.1 & 84.9  & \textcolor{red}{68.7} & 81.2 & 88.4 & 92.7 \\ \hline
MIC+$\mathrm{L}^{\mathcal{W}}$ & \textcolor{red}{70.4} & \textcolor{red}{66.2} & \textcolor{red}{77.4} & 85.2 & 68.3 & 82.2 & 88.9 & 92.9  \\ \Xhline{3\arrayrulewidth}
\end{tabular}
}
\end{table*}

\subsection*{Comparison to State of the Art Deep Metric Learning methods:} In this section, we study the effects of equipping the loss function of the current state-of-the-art \textbf{MIC}~\citep{roth2019mic} algorithm with $\mathrm{L}^{\mathrm{L}}$ and $\mathrm{L}^{\mathcal{W}}$. MIC proposes to use a separate encoder trained with a novel surrogate task to incorporate structure inter-class characteristics while learning a discriminative embedding. The encoder is trained to learn shared characteristics along using standard metric learning loss functions. We have followed the same training protocol as proposed in MIC. We have used ResNet50~\citep{he2016deep} in our implementation of MIC (indicated using $\ast$) and its MMD and Wasserstein counterparts. The various baseline algorithms considered for this study are \textbf{(\RNum{1})} Trip~\citep{Schroff_CVPR_2015}, \textbf{(\RNum{2})} DSC~\citep{Law_ICML_2017}, \textbf{(\RNum{3})} Lifetd-Struct~\citep{Song_CVPR_2016}, \textbf{(\RNum{4})} NPairs~\citep{Sohn_NIPS_2016}, \textbf{(\RNum{5})} NMI-based~\citep{Song_CVPR_2017}, \textbf{(\RNum{6})} Stiefel~\citep{roy2019siamese}, \textbf{(\RNum{7})} Hist~\citep{Ustinova_NIPS_2016}, \textbf{(\RNum{8})} Ang~\citep{wang2017deep}, \textbf{(\RNum{9})} DAML~\citep{duan2018deep}, \textbf{(\RNum{10})} HADML~\citep{zheng2019hardness}, \textbf{(\RNum{11})} DVML~\citep{lin2018deep}, \textbf{(\RNum{12})} BIER~\citep{opitz2017bier}, \textbf{(\RNum{13})} HTL~\citep{ge2018deep}, \textbf{(\RNum{14})} HTG~\citep{zhao2018adversarial}, \textbf{(\RNum{15})} A-BIER~\citep{opitz2018deep}, \textbf{(\RNum{16})} DWS~\citep{Wu_ICCV_2017}, \textbf{(\RNum{17})} ProxyNCA~\citep{movshovitz2017no} and \textbf{(\RNum{18})} MIC~\citep{roth2019mic}. Table~\ref{tab:sota} also shows the results on CUBS-$200$-$2011$~\citep{CUB200_DB} and CARS$196$~\citep{CARS196_DB} where the class-wise information is incorporated into MIC via $\mathrm{L}^{\mathrm{L}}$ and $\mathrm{L}^{\mathcal{W}}$~\footnote{We have repeated the experiments $6$ times with random weights for each trial, and reported the best results according to NMI for $\textrm{MIC}+\mathrm{L}^{\mathrm{L}}$ and \textrm{MIC}$+\mathrm{L}^{\mathcal{W}}$.}. As observed, equipping the loss function of MIC with $\mathrm{L}^{\mathrm{L}}$ and $\mathrm{L}^{\mathcal{W}}$ leads to an increase in terms of NMI and R@K for both datasets. This clearly indicates that MIC also benefits when such comprehensive class-wise constraints are taken into account and a discrepancy loss is enforced between the class-wise embeddings using $\mathrm{L}^{\mathrm{L}}$ and $\mathrm{L}^{\mathcal{W}}$.

\D{In order to validate the performance gain due to $\mathrm{L}^{\mathrm{L}}$ and $\mathrm{L}^{\mathcal{W}}$ over $\textrm{MIC}$, we have performed significance analysis using the Welch t-statistic for unpaired two-samples t-test. The formula for calculating the p-values is as follows} 

\begin{equation}
    \textrm{p} = \frac{\mu_{\textbf{A}} - \mu_{\textbf{B}}}{\sqrt{\frac{\sigma^2_{\textbf{A}}}{N_{\textbf{A}}} + \frac{\sigma^2_{\textbf{B}}}{N_{\textbf{B}}}}} ~~~~,
\end{equation}
\D{where \textbf{A} and \textbf{B} denote the two sets of experiments taken into account. $\mu_{\textbf{A}}$, $\sigma^2_{\textbf{A}}$ and $N_{\textbf{A}}$ denote the sample mean, standard deviation and sample size of the experiment set \textbf{A}. Similar notations for \textbf{B} are represented as $\mu_{\textbf{B}}$, $\sigma^2_{\textbf{B}}$ and $N_{\textbf{B}}$. The sample size (\ie $N_{\textbf{A}}$ and $N_{\textbf{B}}$) is fixed to $6$. \textbf{A} denotes $\textrm{MIC}^\ast$ where as \textbf{B} denotes either $\textrm{MIC}\!+\!\mathrm{L}^{\mathrm{L}}$ and $\textrm{MIC}\!+\!\mathrm{L}^{\mathcal{W}}$.} 
The \say{p-values} between $\textrm{MIC}^\ast$ and $\textrm{MIC}\!+\!\mathrm{L}^{\mathrm{L}}$ are $7\!\times\!10^{-5}$ and $85\!\times\!10^{-6}$ for CUBS-$200$-$2011$ and CARS$196$ datasets respectively. Similarly the p-values between $\textrm{MIC}^\ast$ and $\textrm{MIC}\!+\!\mathrm{L}^{\mathcal{W}}$ are $85\!\times\!10^{-6}$ and $51\!\times\!10^{-3}$ respectively for CUBS-$200$-$2011$ and CARS$196$ datasets. These values clearly indicate that the results obtained are statistically significant over MIC across both the datasets. In the future, one can study the effect of $\mathrm{L}^{\mathrm{L}}$ and $\mathrm{L}^{\mathcal{W}}$ when integrated with other loss functions as mentioned in Table~\ref{tab:sota}. 

\noindent\textbf{Note:} \D{It is to be noted that our proposed algorithm has not been explicitly designed to alleviate the effects of noise from the input data and bridge the performance gap between the noisy and noiseless baseline methods. Indeed, through our detailed empirical results we have demonstrated that our proposed optimal transport based loss function is able to separate the class-wise distributions even in the noisy input data; thus being more robust in handling the side effects of noise compared to the baseline methods.}